\pdfoutput=1

\documentclass[11pt]{article}

\usepackage[]{acl}
\usepackage{multirow}

\usepackage{colortbl}
\usepackage{graphicx}
\usepackage[export]{adjustbox}
\usepackage[normalem]{ulem}
\usepackage{caption}
\usepackage{subcaption}
\usepackage{pifont}
\usepackage{soul}

\usepackage{times}
\usepackage{latexsym}

\usepackage[T1]{fontenc}

\usepackage[utf8]{inputenc}

\usepackage{microtype}

\title{Legal Case Document Summarization: Extractive and Abstractive Methods and their Evaluation}

\author{Abhay Shukla$^{1}$ $\quad$ \textbf{Paheli Bhattacharya}$^{1}$ $\quad$ \textbf{Soham Poddar}$^{1}$ $\quad$ \textbf{Rajdeep Mukherjee}$^{1}$ \\ \textbf{Kripabandhu Ghosh}$^{2}$ $\quad$ \textbf{Pawan Goyal}$^{1}$ $\quad$ \textbf{Saptarshi Ghosh}$^{1}$\thanks{$\;$ Corresponding author: saptarshi@cse.iitkgp.ac.in} \\
        $^{1}$Indian Institute of Technology Kharagpur, India \\ 
        $^{2}$Indian Institute of Science Education and Research Kolkata, India}

\begin{document}
\maketitle
\begin{abstract}
Summarization of legal case judgement documents is a challenging problem in Legal NLP. 
However, not much analyses exist on how different families of summarization models (e.g., extractive vs. abstractive) perform when applied to legal case documents.
This question is particularly important since many recent transformer-based abstractive summarization models have restrictions on the number of input tokens, and legal documents are known to be very long.
Also, it is an open question on how best to evaluate legal case document summarization systems.
In this paper, we carry out extensive experiments with several extractive and abstractive summarization methods (both supervised and unsupervised) over three legal summarization datasets that we have developed. Our analyses, that includes evaluation by law practitioners, lead to several interesting insights on legal summarization in specific and long document summarization in general. 

\end{abstract}

\section{Introduction}
\label{sec:intro}

In Common Law systems (followed in India, UK, USA, etc.) law practitioners have to read through hundreds of case judgements/rulings in order to identify relevant cases that they can cite as precedents in an ongoing case. 
This is a time-consuming process as case documents are generally very long and complex. 
Thus, automatic summarization of legal case documents is an important problem~\cite{gelbart1991beyond,bhattacharya2019comparative,zhong2019automatic,liu2019extracting}. 
It is additionally challenging due to two primary reasons as demonstrated in Table~\ref{tab:related-datasets} -- 
(i)~legal documents as well as their summaries are much longer than most other types of documents, and
(ii)~since it is expensive to get \textit{Law Experts} to write summaries, the datasets are usually much smaller, making it difficult to use supervised models.

\begin{table}[tb]
\centering
\resizebox{\linewidth}{!}{
\begin{tabular}{|c|c|c|c|c|c|}
\hline
\multirow{2}{*}{\textbf{Dataset}} &
\multirow{2}{*}{\textbf{Language}} &
\multirow{2}{*}{\textbf{Domain}} & 
\multirow{2}{*}{\textbf{\#Doc}} & 
\multicolumn{2}{c|}{\textbf{Avg \# Tokens}} \\ 
\cline{5-6} &  &  & & \textbf{Doc} & \textbf{Summ} \\ \hline
CNN/DM~\cite{cnn-dm} & EN & News & 312K & 781 & 56 \\ 
Gigawords~\cite{nyt-gigaword} & EN & News & 4.02M & 31 & 8 \\ 
\hline
arXiv~\cite{arxiv-pubmed} & EN & Academic & 216K & 6,914 & 293 \\ 
PubMed~\cite{arxiv-pubmed} & EN & Academic & 133K & 3,224 & 214 \\ \hline
TL;DR, TOS;DR~\cite{plain-eng-contract} & EN & Contracts & 506 & 106 & 17 \\
BigPatent~\cite{bigpatent} & EN & Patent & 1.34M & 3,573 & 117 \\ 
RulingBR~\cite{rulingbr} & Portugese & Court Rulings & 10,623 & 1,397 & 100 \\ \hline 

\multicolumn{6}{|c|}{\textit{This work}} \\ \hline
IN-Ext (Indian docs, extractive summ) & EN & Court Rulings & 50 & 5,389 & 1,670 \\ 
IN-Abs (Indian docs, abstractive summ) & EN & Court Rulings & 7,130 & 4,378 & 1,051 \\ 
UK-Abs (UK docs, abstractive summ) & EN & Court Rulings & 793 & 14,296 & 1,573 \\ \hline

\end{tabular}
}
\caption{Comparing some existing summarization datasets with the three legal summarization datasets developed in this work. Last two columns give the average number of tokens per document and per summary.
}
\label{tab:related-datasets}
\end{table}


A plethora of solutions exists for text summarization, for e.g., extractive and abstractive, supervised and unsupervised, etc.~\cite{huang-etal-2020-achieved}. 
Also, several legal domain-specific methods have been designed for case document summarization~\cite{zhong2019automatic,liu2019extracting}. 
However, detailed systematic analyses are rare on how the different families of summarization models perform on legal case documents. 
Our prior work~\cite{bhattacharya2019comparative} took an early step in this direction, but it mostly considered extractive methods. 
The state-of-the-art in document summarization has advanced rapidly in the last couple of years, and there has not been much exploration on how recent transformer-based summarization models perform on legal documents~\cite{legalabs,2step}.

To bridge this gap, we 
(1)~develop three legal case judgement summarization datasets from case documents from the Indian and UK Supreme Courts (see Table~\ref{tab:related-datasets}; details in Section~\ref{sec:dataset}), and 
(2)~reproduce/apply representative methods from several families of summarization models on these datasets, and analyse their performances. 
To our knowledge, this is the first study on how a wide spectrum of summarization methods perform over legal case documents. We list below some interesting insights that come out from our analyses.


\noindent $\bullet$ \textbf{Domain-specific vs Domain-agnostic methods}: 
We apply several domain-independent summarization methods, including
\textbf{unsupervised extractive} (e.g., LexRank~\cite{Erkan:2004}, DSDR~\cite{He-dsdr}, and PacSum~\cite{pacsum}),
\textbf{supervised extractive} (e.g., SummaRunner~\cite{nallapati2017summarunner}, and BERTSUMM~\cite{bertsum}), and \textbf{supervised abstractive} (e.g., BART~\cite{bart}, and Longformer~\cite{longformer}) on legal case documents.
We then reproduce several legal domain-specific summarization methods, for e.g.,  
MMR~\cite{zhong2019automatic}, CaseSummarizer~\cite{polsley2016casesummarizer} (unsupervised) and Gist~\cite{liu2019extracting} (supervised).
In many cases, we observe general (domain-agnostic) methods to perform better than domain-specific methods.



\noindent $\bullet$ \textbf{Domain-specific training/fine-tuning}: Using models pretrained on legal corpora, like Legal-Pegasus~\cite{legal-pegasus}, consistently improves performance. 
We also explore and compare multiple ways of generating legal data for training supervised models and further fine-tuning pretrained models. 

\noindent $\bullet$ \textbf{How to deal with long documents}: 
A key challenge in using existing abstractive summarizers on legal documents is that the input capacity of such models is often much lower than the length of legal documents.
Accordingly, we experiment with three different approaches for summarizing long legal case documents --
(i)~applying \textit{long document summarizers} such as Longformer~\cite{longformer} that are designed to handle long documents, 
(ii)~applying \textit{short document summarizers} such as BART~\cite{bart} and Legal-Pegasus~\cite{legal-pegasus} together with approaches for \textit{chunking} the documents, and
(iii)~reducing the size of the input document by first performing an extractive summarization and then going for abstractive summarization. 
In general, we find the chunking-based approach to perform better for legal documents, especially with fine-tuning, although Longformer performs the best on the UK-Abs dataset containing the longest documents, according to some of the metrics.

\noindent $\bullet$ \textbf{Evaluation of summary quality}: 
As noted in~\cite{bhattacharya2019comparative}, \textit{Law Experts} advise to not only evaluate the full-document summaries, but also check how well a summary is able to represent the different logical rhetorical segments in a legal case document (such as Facts, Final Judgement, etc. -- see Appendix, Section~\ref{sec:rhet-roles}). 
To this end, we perform (i)~document-wide automatic evaluations,  (ii)~segment-wise automatic evaluations, as well as (iii)~evaluations by Law practitioners (the actual end-users of legal summarization systems). 

We show that simply computing document-wide metrics gives an incomplete picture of the quality of legal document summarization.
In particular, we see some differences between automatic evaluation and evaluation by domain experts.
For instance, supervised methods like SummaRunner, and finetuned BART usually achieve higher ROUGE scores, but the law practitioners often prefer the summaries generated by simpler unsupervised methods such as DSDR and CaseSummarizer.
Again, the ROUGE scores achieved by the best extractive models are at par with those achieved by the best abstractive models.
However, the practitioners often prefer the extractive summaries over the abstractive ones.

\vspace{1.5 mm}
\noindent \textbf{Availability of resources}:
The three legal summarization datasets curated in this work and the implementations of various summarization models are publicly available at \url{https://github.com/Law-AI/summarization}.


\section{Related Work}
\label{sec:related-work}

We give an overview of existing summarization algorithms~\cite{DBLP:journals/corr/abs-1804-04589,huang-etal-2020-achieved}. 

\vspace{5pt}
\noindent \textbf{Extractive domain-independent methods}: 
There exists a wide range of general/domain-agnostic \textit{unsupervised} summarizers such as Reduction~\cite{jing2000sentence}, and the graph-based LexRank algorithm~\cite{Erkan:2004}. LSA~\cite{Gong:2001} is a matrix-factorization based method and DSDR~\cite{He-dsdr} relies on data reconstruction. PacSum~\cite{pacsum} is a recent BERT-based method.
Among \textit{supervised} neural summarizers, SummaRuNNer~\cite{nallapati2017summarunner} and BERTSum~\cite{bertsum} treat document summarization as a binary classification problem (in-summary vs. out-of-summary). 

\vspace{5pt}
\noindent \textbf{Extractive domain-specific methods}: 
Several domain-specific approaches have been specifically designed for summarizing legal case documents.
Among unsupervised methods, (1)~LetSum~\cite{farzindar2004letsum} and (2)~KMM~\cite{saravanan2006improving} rank sentences based on term distribution models (TF-IDF and k-mixture model respectively);
(3)~CaseSummarizer~\cite{polsley2016casesummarizer} ranks sentences based on their TF-IDF weights coupled with legal-specific features;
(4)~MMR~\cite{zhong2019automatic} generates a template-based summary using a 2-stage classifier and a Maximum Margin Relevance~\cite{zhong2019automatic} module. 

To our knowledge, Gist~\cite{liu2019extracting} is the only supervised method specifically designed for summarizing legal case documents. Gist first represents a sentence with different handcrafted features. It then uses 3 models -- MLP, Gradient Boosted Decision Tree, and LSTM -- to rank sentences in order of their likelihood to be included in the summary. We reproduce all these methods (implementation details in Appendix, Section~\ref{sec:domain-specific-extractive-implementation}). 



\vspace{5pt}
\noindent \textbf{Abstractive methods}:
Most abstractive summarization models have an input token limit which is usually shorter than the length of legal case documents. Approaches from this family include Pointer-Generator~\cite{pointer-generator}, BERTSum-Abs~\cite{bertsum}, Pegasus~\cite{pegasus} and BART~\cite{DBLP:journals/corr/abs-1910-13461} 
(input token limits for these models are at most 1024).
Models like Longformer~\cite{longformer} introduce transformer architectures with more efficient attention mechanisms that enables them to summarize long documents (up to $16 \times 1024$ input tokens). 

~\citet{2step} developed a two-step extractive-abstractive approach for long document summarization -- they use a pre-trained BART model over compressed documents generated by identifying salient sentences. 
In this work, we reproduce a simplified version of this method. 

~\citet{divide&con} presented a divide and conquer approach for long document summarization; they split the documents and summaries, using sentence similarity, into an ensemble of smaller summarization problems. 
In this work, we apply a method inspired by~\citet{divide&con} to fine-tune abstractive models. 

To our knowledge, the only method for abstractive \textit{legal} document summarization is LegalSumm~\cite{legalabs}. 
The method uses the RulingBR dataset (in Portuguese language) which has much shorter documents and summaries than the datasets in this work (see Table~\ref{tab:related-datasets}). 
A limitation of LegalSumm is that it can generate summaries only up to 200 tokens (which is much smaller than our target summaries); hence we do not apply this method in this work.

\section{Datasets for Legal Summarization}
\label{sec:dataset}
There are very few publicly available datasets for legal case document summarization, especially in English (see Table~\ref{tab:related-datasets}). 
In this work, we develop the following three datasets:


\noindent \textbf{(i)~Indian-Abstractive dataset (IN-Abs):} We collect Indian Supreme Court judgements from the website of Legal Information Institute of India (\url{http://www.liiofindia.org/in/cases/cen/INSC/}) which provides free and non-profit access to databases of Indian law. 
Abstractive summaries (also called ``headnotes'') are available for some of these cases; 
of which we include $7,130$ case documents, together with their \textit{headnotes}/summaries as part of the dataset. 
We reserve $100$ randomly-selected document-summary pairs for evaluation and the remaining $7,030$ pairs are used for training the supervised models. 


\vspace{5pt}
\noindent \textbf{(ii)~Indian-Extractive dataset (IN-Ext):} 
Different law practitioners may have different preferences about the summary of a legal case document. 
Per discussion with \textit{Law Experts} (two recent LLB graduates and a Professor from the Rajiv Gandhi School of Intellectual Property Law, a reputed Law school in India), we understand that they are \textit{not} much satisfied with the summaries in the IN-Abs dataset.
According to these experts, legal case documents have various \textit{rhetorical segments}, and the summary should contain a representation from each segment. 
Based on the above preference, the two LLB graduates first rhetorically labelled each sentence from $50$ case documents from the Indian Supreme Court (total 9,380 sentences), with one of the following labels -- \textit{Facts} (abbreviated as FAC), \textit{Argument} (ARG), \textit{Statute} (STA), \textit{Precedent} (PRE), \textit{Ratio of the decision} (Ratio), and \textit{Ruling by Present Court} (RPC). Descriptions of these rhetorical labels are given in the Appendix (Section~\ref{sec:rhet-roles}).
Then they wrote \textit{extractive} summaries for the same $50$ documents, each of length approximately one-third of that of the documents. 
They summarized each rhetorical segment separately; however, they preferred to summarize the segments `Ratio' and `Precedent' together. 
Each LLB graduate was paid a (mutually agreed) honorarium of INR 800 for labeling and summarizing each document.

Since $50$ document-summary pairs are not sufficient for training supervised models, 
when applying these models on IN-Ext, they were trained over the $7,030$ document-summary pairs in the IN-Abs train set. 
We ensure that there is {\it no overlap} between this training set and the IN-Ext dataset. 

\vspace{5pt}
\noindent \textbf{(iii)~UK-Abstractive dataset (UK-Abs):} 
The UK Supreme court website (\url{https://www.supremecourt.uk/decided-cases/}) provides all cases judgements that were ruled since the year 2009. 
For most of the cases, along with the judgements, they also provide the official {\it press summaries} of the cases, which we consider as the reference summary. 
The summaries are abstractive in nature and are divided into three segments -- `Background to the Appeal', `Judgement', and `Reasons for Judgement'. 
We gathered a set of $793$ case documents (decided during the years 2009--2021) and their summaries. 
We reserve $100$ document-summary pairs for evaluation and use the remaining $693$ document-summary pairs for training the supervised models.

\begin{table}[t]
\resizebox{\linewidth}{!}{

\begin{tabular}{|c|c|c|c|c|c|}
\hline
\textbf{Dataset} & \textbf{Type of} &  \textbf{Compression} & \textbf{Test} & \textbf{Training} 
\\ 
 & \textbf{Summary} &  \textbf{Ratio} & \textbf{Set Size} & \textbf{Set Size} 
\\ \hline
IN-Ext & Ext, segmented & 0.31 & 50 & \multirow{2}{*}{7030} \\ \cline{1-4}
IN-Abs & Abs, non-segmented & 0.24 & 100 &  \\ \hline
UK-Abs & Abs, segmented & 0.11  & 100 & 693 \\ \hline
\end{tabular}
}
\caption{The three datasets developed in this work.}
\label{tab:summ-dataset-stats}
\end{table}

\vspace{5pt}
\noindent Table~\ref{tab:summ-dataset-stats} provides a summary of the datasets, while Table~\ref{tab:related-datasets} compares the length of the documents in these datasets with those in other datasets.
Note that the documents in UK-Abs are approximately double the length of the IN-Abs and IN-Ext documents, and have a very low compression ratio (0.11); hence the UK-Abs dataset is the most challenging one for automatic summarization.

\section{Experimental Setup and Evaluation}
\label{sec:exp-setup}


\noindent \textbf{Target length of summaries}: During inference, the trained summarization models need to be provided with the target length of summaries $L$ (in {\it number of words}). 
For every document in the IN-Ext dataset, we have two reference summaries (written by two experts). For a particular document, we consider $L$ to be the average of the number of words in the two reference summaries for that document. 
For IN-Abs and UK-Abs datasets, $L$ is taken as the number of words in the single abstractive reference summary for a given document.

Given a document, every model is made to generate a summary of length at most $L$ words. 
Some algorithms (e.g. KMM, Gist) return a ranking of sentences according to their summary-worthiness. The final summary is obtained by selecting sentences in descending order of the ranked list till the limit of $L$ words is reached.

\vspace{5pt}
\noindent \textbf{Evaluation of summary quality:} 
We report ROUGE-1, ROUGE-2, and ROUGE-L F-scores (computed using \url{https://pypi.org/project/py-rouge/}, with \textit{max\_n} set to 2, parameters \textit{limit\_length} and \textit{length\_limit} not used, and other parameters kept as default), 
and BertScore~\cite{bertscore} (computed using \url{https://pypi.org/project/bert-score/} version 0.3.4) that calculates the semantic similarity scores using the pretrained BERT model. We calculate two kinds of ROUGE and BERTScore as follows:


\vspace{5pt}
\noindent \textit{(a) Overall document-wide scores:} 
For a given document, we compute the ROUGE and BERTScore of an algorithmic summary with respect to the reference summary. 
For IN-Ext, we compute the scores individually with each of the two reference summaries and take the average.  
The scores are averaged over all documents in the evaluation set.

\vspace{5pt}
\noindent \textit{(b) Segment-wise scores:} 
In legal case judgement summarization, a segment-wise evaluation is important to understand how well each rhetorical segment has been summarized~\cite{bhattacharya2019comparative}.
We can perform this evaluation only for the IN-Ext and UK-Abs datasets (and \textit{not} for IN-Abs), where the reference summaries are written segment-wise.
For each rhetorical segment (e.g., \textit{Fact} or \textit{Background}), we extract the portion of the gold standard summary that belongs to that segment. Then we compute the ROUGE score between the entire algorithmic summary and segment-specific part of the reference summary.
We compute the average ROUGE score for a particular segment, averaged over all documents in the evaluation set.\footnote{In this paper, we report segment-wise ROUGE scores only since both segment-wise ROUGE scores as well as segment-wise BERTScores give similar insights.} 


In the segment-wise evaluation, we only report ROUGE Recall scores, and \textit{not}  F-scores. This is because the summarization algorithms output only a coherent set of sentences as summary, and do \textit{not} specify which part of the summary belongs to which segment;  
computing ROUGE Precision or F-Score in this case would be misleading.

\vspace{5pt}
\noindent \textbf{Expert evaluation:} We select a few methods (that achieve the highest ROUGE scores) and get the summaries generated by them for a few documents evaluated by three Law experts (Section~\ref{sec:eval}).

\vspace{5pt}
\noindent \textbf{Consistency scores}: It is important to measure the consistency of an algorithmic summary with the original document, given the possibility of hallucination by abstractive models~\cite{pagnoni2021understanding}. 
To this end, we experimented with the SummaC\textsubscript{CONV} summary consistency checker~\cite{summac}.
However, we find that it gives very low consistency scores to the expert-written reference abstractive summaries -- the average scores for the expert summaries in IN-Abs and UK-Abs are 0.485 and 0.367 respectively. 
A probable reason for these counter-intuitive scores could be that the SummaC\textsubscript{CONV} model could not be fine-tuned on a legal domain-specific dataset, owing to its unavailability. Curating such a dataset to check for factual consistency of summaries of legal documents, together with developing a suitable consistency measure for summaries in the legal domain are envisioned as immediate future works. The present SummaC\textsubscript{CONV} consistency scores are therefore concluded to be unreliable for legal document summarization, and hence are not reported.

\section{Extractive Summarization Methods}
\label{sec:extractive}
\label{sec:extr-impl}

We consider some representative methods from four classes of extractive summarizers:
(1)~Legal domain-specific unsupervised methods: LetSum, KMM, CaseSummarizer, and MMR.
(2)~Legal domain-specific supervised methods: Gist.
(3)~Domain-independent unsupervised methods: LexRank, LSA, DSDR, Luhn, Reduction and PacSum. 
(4)~Domain-independent supervised methods: SummaRuNNer and BERTSum.

Short descriptions of all the above methods are given in Section~\ref{sec:related-work}. The implementation details for the domain-specific methods we implemented, and publicly available code repositories are stated in the Appendix (Section~\ref{sec:domain-specific-extractive-implementation} and Section~\ref{sec:general-extractive-methods-implementation}).


\vspace{5pt}
\noindent \textbf{Training supervised extractive models:}
The supervised methods (Gist, SummaRuNNer and BERTSUM) require labelled training data, where every sentence must be labeled as $1$ if the sentence is suitable for inclusion in the summary, and $0$ otherwise.
As stated in Section~\ref{sec:dataset}, we use  parts of the IN-Abs and UK-Abs datasets for training the supervised methods. 
However, since both these datasets have \textit{abstractive} summaries, they cannot be directly used to train the extractive summarizers. 

We explore three methods -- \textit{Maximal, Avr}, and \textit{TF-IDF} -- for converting the abstractive summaries to their extractive counterparts. 
Best performances for the supervised methods are observed when the training data is generated through the \textbf{Avr} method;
hence we describe \textbf{Avr} here and report results of the supervised methods trained on data generated through \textbf{Avr}. Descriptions of Maximal and TF-IDF are stated in the Appendix (Section~\ref{sec:extractive-supervised-methods-training-data}).


\vspace{5pt}
\noindent \textbf{Avr}: We adopt the technique given by~\citet{narayan2018ranking}. For each sentence in the abstractive gold-standard summary, we select $3$ sentences from the source document (full text) that have the maximum average of ROUGE-1, ROUGE-2 and ROUGE-L scores w.r.t. the sentence in the abstractive summary. 
Then we take the union of all the sentences thus selected, and label them $1$ (to be included in the summary). All other sentences in the source document are assigned a label of $0$. 


\section{Abstractive Summarization Methods}
\label{sec:abstractive}
\label{sec:abstr-impl}

We apply several abstractive methods for legal document summarization, including both pretrained models and models finetuned for legal document summarization. 
A key challenge in applying such methods is that legal documents are usually very long, and most abstractive summarization models have restrictions on the number of input tokens.

\subsection{Pretrained Abstractive Models}
\label{sec:pretr-abs}

\subsubsection{Models meant for short documents}
\label{ss:chunking}

We consider Legal-Pegasus~\cite{legal-pegasus} which is already pretrained on legal documents, and BART~\cite{bart} (max input length of 1024 tokens). 
We use their pre-trained versions from the HuggingFace library; details in the Appendix (Section~\ref{sec:abstractive-implementation}).


The input token limit in these models (1024) is much smaller than the number of words in a typical legal case document.
Hence, to apply these models on legal case documents, we apply a chunking-based approach as described below:



\vspace{5pt}
\noindent \textbf{Chunking-based approach}: 
We first divide a document into small chunks, the size of each chunk being the maximum number of tokens (say, \textit{n}) that a model is designed/pre-trained to accept without truncating (e.g., $n=1024$ for BART). 
Specifically, the first \textit{n} tokens (without breaking sentences) go to the first chunk, the next \textit{n} tokens go to the second chunk, and so on. 
Then we use a model to summarize every chunk. For a given document, we equally divide the target summary length among all the chunks. Finally, we append the generated summaries for each chunk in sequence.

\subsubsection{Models meant for long documents}
\label{ss:long}
Models like Longformer (LED)~\cite{longformer} have been especially designed to handle long documents (input capacity = 16,384 tokens), by including an attention mechanism that scales linearly with sequence length.
We use Legal-LED specifically finetuned on legal data (details in Appendix, Section~\ref{sec:abstractive-implementation}). 
The model could accommodate most case documents fully. A few documents in UK-Abs are however longer (see Table~\ref{tab:summ-dataset-stats}), those documents were truncated after 16,384 tokens.

\subsubsection{Hybrid extractive-abstractive approach}
\label{ss:extabs}

To focus only on important parts of the document in the chunking-based approach,
we use a hybrid of an extractive approach and an abstractive approach, similar to~\citet{2step}. 
First, the document length is reduced by selecting salient sentences using a BERT-based extractive summarization model. 
Then a BART model is used to generate the final summary~\cite{2step}. 
Since, in our case, we often require a summary length greater than 1024 (see Table~\ref{tab:related-datasets}), we use a chunking-based BART (rather than pre-trained BART) in the second step.
We call this model \textbf{BERT\_BART}.



\subsection{Finetuning Abstractive Models}
\label{sec:ft-abs}

Fine-Tuning transformer models has shown significant improvement in most downstream tasks. 
Hence, we finetune BART, Longformer, and Legal-Pegasus on our proposed datasets. 
We also use finetuned BART as part of our BERT\_BART model.  

\vspace{5pt}
\noindent {\bf Generating finetuning data:}
Finetuning supervised models needs a large set of doc-summary pairs. 
However, our considered models (apart from Longformer) have a restricted input limit which is lesser than the length of documents in our datasets. Hence, we use the following method, inspired from~\citet{divide&con}, to generate finetuning data for chunking based summarization.


Consider ($d$, $s$) to be a (training document, reference summary) pair. 
When $d$ is segmented into $n$ chunks $d_1$, $d_2$, ... $d_n$, 
it is not logical for the same $s$ to be the reference summary for each chunk $d_i$. 
In order to generate a suitable reference summary $s_i$ for each chunk $d_i$, first we map every sentence in $s$ to the most similar sentence in $d$. 
Here, we use a variety of sentence-similarity measures, as detailed below.
Then for every chunk $d_i$, we combine all sentences in $s$ which are mapped to any of the sentences in $d_i$, and consider those sentences as the summary $s_i$ (of $d_i$). 
Following this procedure, from each document, we get a large number of ($d_i$, $s_i$) pairs which are then used for finetuning. 

\vspace{5pt}
\noindent {\bf Sentence similarity measures for generating fine-tuning data:} 
We experiment with several techniques for measuring sentence similarity between two sentences -- (i)~Mean Cosine Similarity (\textbf{MCS}), (ii)~Smooth Inverse Frequency (\textbf{SIF}), (iii)~Cosine similarity between BERT [CLS] token embeddings (\textbf{CLS}), and (iv)~\textbf{MCS\_RR} which incorporates rhetorical role information. 
Out of these, we find MCS to perform the best. 
Hence we describe MCS in detail here. Descriptions of the other methods can be found in the Appendix (Section~\ref{sec:abstractive-finetuning-methods}).

In Mean Cosine Similarity (\textbf{MCS})~\cite{ranasinghe-etal-2019-enhancing}, we calculate the mean of token-level embeddings (obtained using SBERT~\cite{reimers-2019-sentence-bert}) to obtain the representation for a given sentence. 
We then compute the cosine similarity between two such sentence embeddings.

We used all the methods stated above to generate fine-tuning datasets for IN-Abs and UK-Abs. We finetune three different versions of the BART model, BART\_CLS, BART\_MCS, and BART\_SIF, using the three sentence similarity measures described above.
Out of these, BART\_MCS performs the best (as we will see in Section~\ref{sec:results}).
Therefore, we use MCS for generating finetuning data for the other models, to obtain Legal-Pegasus-MCS and BART\_MCS\_RR (where the finetuning data is generated based on rhetorical labels). 
We also use the finetuned BART\_MCS model with BERT\_BART method to get BERT\_BART\_MCS. 

The hyper-parameters used to finetune the different abstractive models are stated in  Table~\ref{tab:hyp} in the Appendix (Section~\ref{sec:abstractive-implementation}).

 
 
 
 




\section{Results and Analyses}
\label{sec:results}
This section analyzes the performance of different summarization models. 
For IN-Ext, In-Abs and UK-Abs datasets,  Table~\ref{tab:results-overall-in-ext}, Table~\ref{tab:results-overall-in-abs} and Table~\ref{tab:results-overall-uk-abs} report the overall evaluation of a few of the  best-performing methods, respectively. Table~\ref{tab:rouge-segment-india} and Table~\ref{tab:rouge-segment-uk} show the segment-wise evaluation of a few best-performing methods on the IN-Ext and UK-Abs datasets respectively. Detailed results are given in Tables~\ref{tab:results-overall-in-ext-all}--\ref{tab:rouge-segment-uk-all} in the Appendix (Section~\ref{sec:detailed-results}). 



\begin{table}[t]
\centering
\resizebox{\linewidth}{!}{
\begin{tabular}{|c|c|c|c|c|c|}
\hline
\multicolumn{1}{|c|}{\multirow{2}{*}{\textbf{Algorithm}}} & \multicolumn{3}{c|}{\textbf{ROUGE Scores}}                                                                      & \multirow{2}{*}{\textbf{BERTScore}} \\ \cline{2-4}
\multicolumn{1}{|c|}{}                                    & \multicolumn{1}{c|}{R-1}            & \multicolumn{1}{c|}{R-2}            & \multicolumn{1}{c|}{R-L}            &                                     \\ \hline
\multicolumn{5}{|c|}{\textit{Extractive Methods}}                                                                                                                                                                 \\ \hline
\multicolumn{5}{|c|}{Unsupervised, Domain Independent}                                                                                                                                                            \\ \hline
\multicolumn{1}{|c|}{Luhn}                                & \multicolumn{1}{c|}{0.568}          & \multicolumn{1}{c|}{0.373}          & \multicolumn{1}{c|}{\textbf{0.422}} & \textbf{0.882}                      \\ \hline
\multicolumn{1}{|c|}{Pacsum\_bert}                        & \multicolumn{1}{c|}{\textbf{0.59}}  & \multicolumn{1}{c|}{\textbf{0.41}}  & \multicolumn{1}{c|}{0.335}          & 0.879                               \\ \hline
\multicolumn{5}{|c|}{Unsupervised, Legal Domain Specific}                                                                                                                                                         \\ \hline
\multicolumn{1}{|c|}{MMR}                                 & \multicolumn{1}{c|}{0.563}          & \multicolumn{1}{c|}{0.318}          & \multicolumn{1}{c|}{0.262}          & 0.833                               \\ \hline
\multicolumn{1}{|c|}{KMM}                      & \multicolumn{1}{c|}{0.532}          & \multicolumn{1}{c|}{0.302}          & \multicolumn{1}{c|}{0.28}           & 0.836                               \\ \hline
\multicolumn{1}{|c|}{LetSum}                              & \multicolumn{1}{c|}{\textbf{0.591}} & \multicolumn{1}{c|}{\textbf{0.401}} & \multicolumn{1}{c|}{\textbf{0.391}} & \textbf{0.875}                      \\ \hline
\multicolumn{5}{|c|}{Supervised, Domain Independent}                                                                                                                                                              \\ \hline
\multicolumn{1}{|c|}{SummaRunner}                         & \multicolumn{1}{c|}{0.532}          & \multicolumn{1}{c|}{0.334}          & \multicolumn{1}{c|}{0.269}          & 0.829                               \\ \hline
\multicolumn{1}{|c|}{BERT-Ext}                            & \multicolumn{1}{c|}{\textbf{0.589}} & \multicolumn{1}{c|}{\textbf{0.398}} & \multicolumn{1}{c|}{\textbf{0.292}} & \textbf{0.85}                       \\ \hline
\multicolumn{5}{|c|}{Supervised, Legal Domain Specific}                                                                                                                                                           \\ \hline
\multicolumn{1}{|c|}{Gist}                                & \multicolumn{1}{c|}{0.555}          & \multicolumn{1}{c|}{0.335}          & \multicolumn{1}{c|}{0.391}          & 0.864                               \\ \hline
\multicolumn{5}{|c|}{\textit{Abstractive Methods}}                                                                                                                                                                \\ \hline
\multicolumn{5}{|c|}{Pretrained}                                                                                                                                                                                  \\ \hline
\multicolumn{1}{|c|}{BART}                                & \multicolumn{1}{c|}{0.475}          & \multicolumn{1}{c|}{0.221}          & \multicolumn{1}{c|}{0.271}          & 0.833                               \\ \hline
\multicolumn{1}{|c|}{BERT-BART}                           & \multicolumn{1}{c|}{\textbf{0.488}} & \multicolumn{1}{c|}{\textbf{0.236}} & \multicolumn{1}{c|}{\textbf{0.279}} & 0.836                               \\ \hline
\multicolumn{1}{|c|}{Legal-Pegasus}                             & \multicolumn{1}{c|}{0.465}          & \multicolumn{1}{c|}{0.211}          & \multicolumn{1}{c|}{\textbf{0.279}} & \textbf{0.842}                      \\ \hline
\multicolumn{1}{|c|}{Legal-LED}                          & \multicolumn{1}{c|}{0.175}          & \multicolumn{1}{c|}{0.036}          & \multicolumn{1}{c|}{0.12}           & 0.799                               \\ \hline
\multicolumn{5}{|c|}{Finetuned}                                                                                                                                                                    \\ \hline
\multicolumn{1}{|c|}{BART\_MCS}                           & \multicolumn{1}{c|}{0.557}          & \multicolumn{1}{c|}{0.322}          & \multicolumn{1}{c|}{0.404}          & 0.868                               \\ \hline
\multicolumn{1}{|c|}{BART\_MCS\_RR}                       & \multicolumn{1}{c|}{0.574}          & \multicolumn{1}{c|}{0.345}          & \multicolumn{1}{c|}{0.402}          & 0.864                               \\ \hline

\multicolumn{1}{|c|}{BERT\_BART\_MCS}                     & \multicolumn{1}{c|}{0.553}          & \multicolumn{1}{c|}{0.316}          & \multicolumn{1}{c|}{0.403}          & \textbf{0.869}                      \\ \hline

\multicolumn{1}{|c|}{Legal-Pegasus\_MCS}                        & \multicolumn{1}{c|}{\textbf{0.575}} & \multicolumn{1}{c|}{\textbf{0.351}} & \multicolumn{1}{c|}{\textbf{0.419}} & 0.864                               \\ \hline
\multicolumn{1}{|c|}{Legal-LED}                          & \multicolumn{1}{c|}{0.471}          & \multicolumn{1}{c|}{0.26}           & \multicolumn{1}{c|}{0.341}          & 0.863                               \\ \hline
\end{tabular}
}
\caption{Document-wide ROUGE-L and BERTScores (FScore) on the IN-Ext dataset. All values averaged over the $50$ documents in the dataset. The best value in a particular class of methods is highlighted in \textbf{bold}.}
\vspace{15pt}
\label{tab:results-overall-in-ext}
\end{table}

\begin{table}[t]
\centering
\resizebox{\linewidth}{!}{
\begin{tabular}{|c|c|c|c|c|}
\hline
\multicolumn{1}{|c|}{\multirow{2}{*}{\textbf{Algorithm}}} & \multicolumn{3}{c|}{\textbf{ROUGE Scores}}                                                                      & \multirow{2}{*}{\textbf{BERTScore}} \\ \cline{2-4}
\multicolumn{1}{|c|}{}                                    & \multicolumn{1}{c|}{R-1}            & \multicolumn{1}{c|}{R-2}            & \multicolumn{1}{c|}{R-L}            &                                     \\ \hline
\multicolumn{5}{|c|}{\textit{Extractive Methods (U: Unsupervised, S: Supervised)}}                                                                                                                                                                 \\ \hline
\multicolumn{1}{|c|}{DSDR (U)}                                & 0.485 & 0.222 & 0.270           & 0.848                      \\ \hline
\multicolumn{1}{|c|}{CaseSummarizer (U)}                      & 0.454 & 0.229 & 0.279 & 0.843                               \\ \hline
\multicolumn{1}{|c|}{SummaRunner (S)}                         & \multicolumn{1}{c|}{\textbf{0.493}} & \multicolumn{1}{c|}{\textbf{0.255}} & 0.274 & \textbf{0.849}                      \\ \hline
\multicolumn{1}{|c|}{Gist (S)}                                & \multicolumn{1}{c|}{0.471}          & \multicolumn{1}{c|}{0.238}          & {\bf 0.308}          & 0.842                               \\ \hline
\multicolumn{5}{|c|}{\textit{Finetuned Abstractive Methods}}                                                                                                                                                                \\ \hline
\multicolumn{1}{|c|}{BART\_MCS}                           & \multicolumn{1}{c|}{\textbf{0.495}} & \multicolumn{1}{c|}{0.249}          & \multicolumn{1}{c|}{0.330}           & 0.851                               \\ \hline
\multicolumn{1}{|c|}{BERT\_BART\_MCS}                     & \multicolumn{1}{c|}{0.487}          & \multicolumn{1}{c|}{0.243}          & \multicolumn{1}{c|}{0.329}          & 0.853                               \\ \hline
\multicolumn{1}{|c|}{Legal-Pegasus\_MCS}                         & \multicolumn{1}{c|}{0.488}          & \multicolumn{1}{c|}{\textbf{0.252}} & \multicolumn{1}{c|}{\textbf{0.341}} & 0.851                               \\ \hline
\multicolumn{1}{|c|}{Legal-LED}                          & \multicolumn{1}{c|}{0.471}          & \multicolumn{1}{c|}{0.235}          & \multicolumn{1}{c|}{0.332}          & \textbf{0.856}                      \\ \hline

\end{tabular}
}
\caption{Document-wide ROUGE-L and BERTScores (Fscore) on the IN-Abs dataset, averaged over the $100$ test documents. Results of some of the top-performing methods are shown here (all results in Table~\ref{tab:results-overall-in-abs-all}).}
\label{tab:results-overall-in-abs}
\end{table}

\begin{table}[t]
\centering
\resizebox{\linewidth}{!}{
\begin{tabular}{|ccccc|}
\hline
\multicolumn{1}{|c|}{\multirow{2}{*}{\textbf{Algorithm}}} & \multicolumn{3}{c|}{\textbf{ROUGE Scores}}                                                                      & \multirow{2}{*}{\textbf{BERTScore}} \\ \cline{2-4}
\multicolumn{1}{|c|}{}                                    & \multicolumn{1}{c|}{R-1}            & \multicolumn{1}{c|}{R-2}            & \multicolumn{1}{c|}{R-L}            &                                     \\ \hline
\multicolumn{5}{|c|}{\textit{Extractive Methods (U: Unsupervised, S: Supervised)}}                                                                                                                                                                 \\ \hline
\multicolumn{1}{|c|}{DSDR (U)}                                & \multicolumn{1}{c|}{0.484} & \multicolumn{1}{c|}{0.174}          & \multicolumn{1}{c|}{0.221}          & 0.832                               \\ \hline
\multicolumn{1}{|c|}{CaseSummarizer (U)}                      & \multicolumn{1}{c|}{0.445} & \multicolumn{1}{c|}{0.166} & \multicolumn{1}{c|}{0.227}          & 0.835                               \\ \hline
\multicolumn{1}{|c|}{SummaRunner (S)}                         & \multicolumn{1}{c|}{\textbf{0.502}} & \multicolumn{1}{c|}{\textbf{0.205}} & \multicolumn{1}{c|}{\textbf{0.237}} & \textbf{0.846}                      \\ \hline
\multicolumn{1}{|c|}{Gist}                                & \multicolumn{1}{c|}{0.427}          & \multicolumn{1}{c|}{0.132}          & \multicolumn{1}{c|}{0.215}          & 0.819                               \\ \hline
\multicolumn{5}{|c|}{\textit{Finetuned Abstractive Methods}}                                                                                                                                                                \\ \hline
\multicolumn{1}{|c|}{BART\_MCS}                           & \multicolumn{1}{c|}{\textbf{0.496}} & \multicolumn{1}{c|}{\textbf{0.188}} & \multicolumn{1}{c|}{\bf 0.271}          & \textbf{0.848}                      \\ \hline
\multicolumn{1}{|c|}{BERT\_BART\_MCS}                     & \multicolumn{1}{c|}{0.476}          & \multicolumn{1}{c|}{0.172}          & \multicolumn{1}{c|}{0.259}          & 0.847                               \\ \hline
\multicolumn{1}{|c|}{Legal-Pegasus\_MCS}                         & \multicolumn{1}{c|}{0.476}          & \multicolumn{1}{c|}{0.171}          & \multicolumn{1}{c|}{0.261}          & 0.838                               \\ \hline
\multicolumn{1}{|c|}{Legal-LED}                          & \multicolumn{1}{c|}{0.482}          & \multicolumn{1}{c|}{0.186}          & \multicolumn{1}{c|}{0.264}          & 0.851                               \\ \hline

\end{tabular}
}
\caption{Document-wide ROUGE-L and BERTScores (Fscore) on UK-Abs dataset, averaged over the $100$ test documents. Results of some of the top-performing methods are shown here (all results in Table~\ref{tab:results-overall-uk-abs-all}).}
\vspace{10pt}
\label{tab:results-overall-uk-abs}

\end{table}

\begin{table}[t]
\resizebox{\linewidth}{!}{
\begin{tabular}{|cccccc|}
\hline
\multicolumn{1}{|c|}{\multirow{2}{*}{\textbf{Algorithms}}} &  \multicolumn{5}{c|}{\textbf{Rouge L Recall}} \\ \cline{2-6} 
\multicolumn{1}{|c|}{} & \multicolumn{1}{c|}{\begin{tabular}[c]{@{}c@{}}\textbf{RPC}\\ (6.42\%)\end{tabular}} &  \multicolumn{1}{c|}{\begin{tabular}[c]{@{}c@{}}\textbf{FAC}\\ (34.85\%)\end{tabular}} &  \multicolumn{1}{c|}{\begin{tabular}[c]{@{}c@{}}\textbf{STA}\\ (13.42\%)\end{tabular}} & \multicolumn{1}{c|}{\begin{tabular}[c]{@{}c@{}}\textbf{Ratio+Pre}\\ (28.83\%)\end{tabular}} &  \multicolumn{1}{c|}{\begin{tabular}[c]{@{}c@{}}\textbf{ARG}\\ (16.45\%)\end{tabular}} \\ \hline
\multicolumn{6}{|c|}{\textit{Extractive Methods (U: Unsupervised, S: Supervised)}}                                                                                                                                                                                  \\ \hline
\multicolumn{1}{|c|}{LexRank (U)}                             & \multicolumn{1}{c|}{0.039}          & \multicolumn{1}{c|}{0.204}          & \multicolumn{1}{c|}{0.104}          & \multicolumn{1}{c|}{0.208}          & \textbf{0.127} \\ \hline
\multicolumn{1}{|c|}{Luhn (U)}                                & \multicolumn{1}{c|}{0.037}          & \multicolumn{1}{c|}{\textbf{0.272}} & \multicolumn{1}{c|}{0.097}          & \multicolumn{1}{c|}{0.175}          & 0.117          \\ \hline
\multicolumn{1}{|c|}{LetSum (U)}                              & \multicolumn{1}{c|}{0.036}          & \multicolumn{1}{c|}{0.237}          & \multicolumn{1}{c|}{\textbf{0.115}} & \multicolumn{1}{c|}{0.189}          & 0.1            \\ \hline
\multicolumn{1}{|c|}{SummaRunner (S)}                         & \multicolumn{1}{c|}{\textbf{0.059}} & \multicolumn{1}{c|}{0.158}          & \multicolumn{1}{c|}{0.08}           & \multicolumn{1}{c|}{0.209}          & 0.096          \\ \hline
\multicolumn{1}{|c|}{Gist (S)}                                & \multicolumn{1}{c|}{0.041}          & \multicolumn{1}{c|}{0.191}          & \multicolumn{1}{c|}{0.102}          & \multicolumn{1}{c|}{\textbf{0.223}} & 0.093          \\ \hline
\multicolumn{6}{|c|}{\textit{Finetuned Abstractive Methods}}                                                                                                                                                                       \\ \hline
\multicolumn{1}{|c|}{BART\_MCS\_RR}                       & \multicolumn{1}{c|}{\textbf{0.061}} & \multicolumn{1}{c|}{0.192}          & \multicolumn{1}{c|}{0.082}          & \multicolumn{1}{c|}{0.237}          & 0.086          \\ \hline
\multicolumn{1}{|c|}{Legal-Pegasus\_MCS}                        & \multicolumn{1}{c|}{0.037}          & \multicolumn{1}{c|}{0.192}          & \multicolumn{1}{c|}{\textbf{0.09}}  & \multicolumn{1}{c|}{\textbf{0.257}} & 0.101 \\ \hline
\multicolumn{1}{|c|}{Legal-LED}                          & \multicolumn{1}{c|}{0.053}          & \multicolumn{1}{c|}{\textbf{0.245}} & \multicolumn{1}{c|}{0.086}          & \multicolumn{1}{c|}{0.187}          & \textbf{0.124}          \\ \hline

\end{tabular}
}
\caption{Segment-wise ROUGE-L Recall scores of the best methods in Table~\ref{tab:results-overall-in-ext} on the IN-Ext dataset. All values are averaged over the $50$ documents in the dataset. The best scores for each segment in a particular class of methods are in \textbf{bold}. Results of all methods in Table~\ref{tab:rouge-segment-india-all}.}
\vspace{-5pt}
\label{tab:rouge-segment-india}
\end{table}

\begin{table}[t]
\resizebox{\linewidth}{!}{
\begin{tabular}{|c|c|c|c|}
\hline
\multicolumn{1}{|c|}{\multirow{2}{*}{\textbf{Algorithms}}}  & \multicolumn{3}{c|}{\textbf{Rouge-L Recall}} \\ \cline{2-4}
 & \multicolumn{1}{c|}{\begin{tabular}[c]{@{}c@{}}\textbf{Background}\\ (39\%)\end{tabular}} & \multicolumn{1}{c|}{\begin{tabular}[c]{@{}c@{}}\textbf{Final Judgement}\\ (5\%)\end{tabular}} & \begin{tabular}[c]{@{}c@{}}\textbf{Reasons}\\ (56\%)\end{tabular} \\ \hline

\multicolumn{4}{|c|}{\textit{Extractive Methods (U: Unsupervised, S: Supervised)}}                                                                                                       \\ \hline
\multicolumn{1}{|c|}{SummaRunner (S)}                          & \multicolumn{1}{c|}{0.172}          & \multicolumn{1}{c|}{\textbf{0.044}} & 0.165          \\ \hline
\multicolumn{1}{|c|}{BERT-Ext (S)}                             & \multicolumn{1}{c|}{\textbf{0.203}} & \multicolumn{1}{c|}{0.034}          & 0.135          \\ \hline
\multicolumn{1}{|c|}{Gist (S)}                                 & \multicolumn{1}{c|}{0.123}          & \multicolumn{1}{c|}{0.041}          & \textbf{0.195} \\ \hline
\multicolumn{4}{|c|}{\textit{Finetuned Abstractive Methods}}                                                                                            \\ \hline
\multicolumn{1}{|c|}{Legal-Pegasus\_MCS}                              & \multicolumn{1}{c|}{0.166}          & \multicolumn{1}{c|}{0.039}          & \textbf{0.202} \\ \hline
\multicolumn{1}{|c|}{Legal-LED}                           & \multicolumn{1}{c|}{\textbf{0.187}} & \multicolumn{1}{c|}{\textbf{0.058}} & 0.172          \\ \hline

\end{tabular}
}
\caption{Segment-wise ROUGE-L Recall scores of the best methods in Table~\ref{tab:results-overall-uk-abs} on the UK-Abs dataset. All values averaged over the $100$ documents in the evaluation set. Best scores for each segment in a particular class of methods are in \textbf{bold}. Results of all methods in Table~\ref{tab:rouge-segment-uk-all}.}
\vspace{10pt}
\label{tab:rouge-segment-uk}

\end{table}

\subsection{Evaluation of Extractive methods}

\textbf{Overall Evaluation (Tables~\ref{tab:results-overall-in-ext}--\ref{tab:results-overall-uk-abs}):} 
Among the unsupervised general methods, Luhn (on IN-Ext) and DSDR (on IN-Abs and UK-Abs) show the best performances.
Among the unsupervised legal-specific methods, CaseSummarizer performs the best on both In-Abs and UK-Abs datasets, while LetSum performs the best on IN-Ext. 
Among supervised extractive methods, SummaRuNNer performs the best across both domain-independent and domain-specific categories, on the IN-Abs and UK-Abs datasets. BERT-Ext is the best performing model on the IN-Ext dataset.

\vspace{2pt}
\noindent \textbf{Segment-wise Evaluation:}
Table~\ref{tab:rouge-segment-india} and Table~\ref{tab:rouge-segment-uk} show the segment-wise ROUGE-L Recall scores of some of the best performing methods on the IN-Ext and UK-Abs datasets respectively. 
Section~\ref{sec:exp-setup} details the process of obtaining these scores. 
According to overall ROUGE scores, it may seem that a particular method performs very well (e.g., LetSum on In-Ext), but that method \textit{may not perform the best across all the segments} (e.g. among the extractive methods, LetSum performs the best in only 1 out of the 5 segments in In-Ext). 
This observation shows the importance of segment-wise evaluation. 
It is an open challenge to develop an algorithm that shows a balanced segment-wise performance. 
Some more interesting observations on segment-wise evaluations are given in the Appendix (Section~\ref{sec:segment-wise-eval-detailed-insights}).

\begin{table*}[t]
\centering
\resizebox{\linewidth}{!}{
\begin{tabular}{|c|cc|cc|cc|cc|cc|cc|cc|cc|cc|}
\hline
\multirow{2}{*}{\textbf{Algorithms}} & \multicolumn{2}{c|}{\textbf{RPC}} & \multicolumn{2}{c|}{\textbf{FAC}} & \multicolumn{2}{c|}{\textbf{STA}} & \multicolumn{2}{c|}{\textbf{PRE}} & \multicolumn{2}{c|}{\textbf{Ratio}} & \multicolumn{2}{c|}{\textbf{ARG}} & \multicolumn{2}{c|}{\textbf{Imp.Inf.}} & \multicolumn{2}{c|}{\textbf{Read.}} & \multicolumn{2}{c|}{\textbf{Overall}} \\ \cline{2-19} 
 & \multicolumn{1}{c|}{Mean} & Med. & \multicolumn{1}{c|}{Mean} & Med. & \multicolumn{1}{c|}{Mean} & Med. & \multicolumn{1}{c|}{Mean} & Med. & \multicolumn{1}{c|}{Mean} & Med. & \multicolumn{1}{c|}{Mean} & Med. & \multicolumn{1}{c|}{Mean} & Med. & \multicolumn{1}{c|}{Mean} & Med. & \multicolumn{1}{c|}{Mean} & Med. \\ \hline
\textbf{DSDR} & \multicolumn{1}{c|}{4.2} & 5 & \multicolumn{1}{c|}{3.8} & 4 & \multicolumn{1}{c|}{3.7} & 4 & \multicolumn{1}{c|}{3.1} & 3.7 & \multicolumn{1}{c|}{3.7} & 4 & \multicolumn{1}{c|}{1.9} & 3.7 & \multicolumn{1}{c|}{3.7} & 4 & \multicolumn{1}{c|}{4.3} & 4 & \multicolumn{1}{c|}{3.9} & 4 \\ \hline
\textbf{CaseSummarizer} & \multicolumn{1}{c|}{2.1} & 2 & \multicolumn{1}{c|}{3.8} & 4 & \multicolumn{1}{c|}{3.6} & 4 & \multicolumn{1}{c|}{3} & 3.6 & \multicolumn{1}{c|}{3.5} & 3 & \multicolumn{1}{c|}{2.4} & 3 & \multicolumn{1}{c|}{3.2} & 3 & \multicolumn{1}{c|}{4.3} & 4 & \multicolumn{1}{c|}{3.6} & 3 \\ \hline
\textbf{SummaRuNNer} & \multicolumn{1}{c|}{2.1} & 3 & \multicolumn{1}{c|}{4.2} & 4 & \multicolumn{1}{c|}{2.4} & 3 & \multicolumn{1}{c|}{3.3} & 3 & \multicolumn{1}{c|}{2.9} & 3 & \multicolumn{1}{c|}{2.1} & 2.9 & \multicolumn{1}{c|}{3.2} & 3 & \multicolumn{1}{c|}{4.1} & 4 & \multicolumn{1}{c|}{3.2} & 4 \\ \hline
\textbf{Gist} & \multicolumn{1}{c|}{3.3} & 4 & \multicolumn{1}{c|}{1.8} & 3 & \multicolumn{1}{c|}{2.6} & 3 & \multicolumn{1}{c|}{3.5} & 3 & \multicolumn{1}{c|}{3.2} & 4 & \multicolumn{1}{c|}{2.1} & 3.2 & \multicolumn{1}{c|}{3} & 3 & \multicolumn{1}{c|}{3.9} & 4 & \multicolumn{1}{c|}{3.2} & 3 \\ \hline
\textbf{Legal-Pegasus} & \multicolumn{1}{c|}{1.4} & 1 & \multicolumn{1}{c|}{3.9} & 4 & \multicolumn{1}{c|}{3.2} & 4 & \multicolumn{1}{c|}{2.4} & 3.2 & \multicolumn{1}{c|}{2.9} & 3 & \multicolumn{1}{c|}{2} & 2.9 & \multicolumn{1}{c|}{3} & 3 & \multicolumn{1}{c|}{3.5} & 4 & \multicolumn{1}{c|}{3} & 3 \\ \hline
\textbf{BART-MCS} & \multicolumn{1}{c|}{0.9} & 1 & \multicolumn{1}{c|}{2.8} & 3 & \multicolumn{1}{c|}{2.9} & 3 & \multicolumn{1}{c|}{3.3} & 3 & \multicolumn{1}{c|}{2.5} & 3 & \multicolumn{1}{c|}{1.8} & 2.5 & \multicolumn{1}{c|}{2.8} & 3 & \multicolumn{1}{c|}{2.7} & 3 & \multicolumn{1}{c|}{2.8} & 3 \\ \hline
\textbf{BART-MCS-RR} & \multicolumn{1}{c|}{0.8} & 1 & \multicolumn{1}{c|}{2.7} & 3 & \multicolumn{1}{c|}{3.1} & 3 & \multicolumn{1}{c|}{2.6} & 3 & \multicolumn{1}{c|}{2.6} & 3 & \multicolumn{1}{c|}{1.3} & 2.6 & \multicolumn{1}{c|}{2.6} & 3 & \multicolumn{1}{c|}{2.9} & 3 & \multicolumn{1}{c|}{2.6} & 3 \\ \hline
\end{tabular}
}
\caption{Evaluation of some summaries from the IN-Abs dataset, by three domain experts (two recent LLB graduates and a Senior faculty of Law). The evaluation parameters are explained in the text. Scores are given by each expert in the range [0-5], $5$ being the best. The Mean and Median (Med.) scores for each summarization algorithm and for each parameter are computed over $15$ scores (across $5$ documents;  each judged by $3$ experts).}
\label{tab:expert-summ-eval}
\vspace*{-10pt}
\end{table*}

\subsection{Evaluation of Abstractive methods}

\noindent \textbf{Overall Evaluation (Tables~\ref{tab:results-overall-in-ext}--\ref{tab:results-overall-uk-abs}):} Among the pretrained models, Legal-Pegasus generates the best summaries (Table~\ref{tab:results-overall-in-ext}), followed by BART-based methods. This is expected, since Legal-Pegasus is pre-trained on legal documents. This short document summarizer, when used with chunking to handle long documents, notably outperforms Legal-LED, which is meant for long documents. 
For IN-Ext dataset, BERT\_BART performs the best maybe due to extractive nature of the summaries.


All models show notable improvement through fine-tuning. Overall, the best performances are noted by Legal-Pegasus (IN-Ext and IN-Abs) and BART\_MCS (UK-Abs).





\noindent \textbf{Segment-wise Evaluation (Tables~\ref{tab:rouge-segment-india}, \ref{tab:rouge-segment-uk}):} Again, none of the methods performs well across all segments, and fine-tuning generally improves performance. 
Interestingly, though Legal-LED performs poorly with respect to document-wide ROUGE scores, it shows better performance in segment-wise evaluation -- it gives the best performance in the FAC and ARG segments of IN-Ext and in 2 out of the 3 segments of UK-Abs.
Since the UK-Abs dataset contains the longest documents, possibly Legal-LED has an advantage over chunking-based methods when evaluated segment-wise.




\vspace{5pt}
\noindent \textbf{Overall performance on long legal case documents}: 
We experimented with three approaches for summarizing long documents -- (i)~models with modified attention mechanism such as Legal-LED, 
(ii)~methods based on chunking the documents, and
(iii)~reducing the size of the input by initial extractive summarization and then going for abstractive summarization (BERT\_BART). 
When we see the overall (document-wide) ROUGE scores,  \textbf{Legal-Pegasus} and \textbf{BART} (when used along with chunking), are seen to perform the best, followed by BERT\_BART.
However for segment-wise performances \textbf{Legal-LED} shows greater potential.

\subsection{Expert evaluation}
\label{sec:eval}

Finally, we evaluate some of the model-generated summaries via three domain experts. Since it is expensive to obtain evaluations from Law experts, we chose to conduct this evaluation for a few documents/summaries from the IN-Abs dataset.

\vspace{1mm}
\noindent 
{\bf Recruiting the 3 experts:}
We recruited the two recent LLB graduates (who wrote the reference summaries in IN-Ext) from the Rajiv Gandhi School of Intellectual Property Law (RGSOIPL), India, who were mentored by a Professor of the same Law school (as mentioned in Section~\ref{sec:dataset}) while carrying out the annotations. 
Additionally, we recruited a senior Faculty of Law from the West Bengal National University of Juridical Sciences (WBNUJS), India. 
Note that both RGSOIPL and WBNUJS are among the most reputed Law schools in India. 

Each annotator was paid a (mutually agreed) honorarium of INR 200 for evaluation of each summary. 
The annotators were clearly informed of the purpose of the survey.
Also we discussed their experiences after the survey about. Through all these steps, we tried our best to ensure that the annotations were done rigorously.

\vspace{1mm}
\noindent {\bf Survey setup:} 
We select the summaries generated by 7 algorithms which give relatively high ROUGE-L F-Score on IN-Abs -- see Table~\ref{tab:expert-summ-eval}. 
Then, we show the annotators 5 selected documents and their summaries generated by the 7 algorithms (35 summaries evaluated in total). 
An annotator was asked to evaluate a summary on the basis of the following parameters  --
\textbf{(1)}~how well a summary represents each rhetorical segment, 
i.e., the final judgement~(\textbf{RPC}), facts~(\textbf{FAC}), relevant statutes/laws cited~(\textbf{STA}), relevant precedents cited~(\textbf{PRE}), the reasoning/rationale behind the judgement~(\textbf{Ratio}), and the arguments presented in the case~(\textbf{ARG}). 
\textbf{(2)}~how well important information has been covered in the summary (\textbf{Imp Inf}). 
\textbf{(3)}~Readability and grammatical coherence (\textbf{Read}). 
\textbf{(4)}~An overall score for the summary (\textbf{Overall}).

Each summary was rated on a Likert scale of $0-5$ on each parameter, independently by the 3 annotators. 
Thus, a particular method got $15$ scores for each parameter -- for 5 documents and by 3 annotators. Table~\ref{tab:expert-summ-eval} reports (i)~the mean/average, and (ii)~the median of all these 15 scores for each method and for each parameter. 


\vspace{2mm}
\noindent{\bf Inter-Annotator Agreement}: We calculate pair-wise Pearson Correlation between the `Overall' scores given by the three annotators over the 35 summaries, and then take the average correlation value as the IAA.
Refer to the Appendix (Section~\ref{sec:expert-eval-details}) for why we chose this IAA measure.
The average IAA is $0.525$ which shows moderate agreement between the annotators\footnote{\url{https://www.andrews.edu/~calkins/math/edrm611/edrm05.htm}}.

\vspace{2mm}
\noindent {\bf Results (Table~\ref{tab:expert-summ-eval}):}
According to the Law experts, important information (Imp. Inf.) could be covered best by DSDR, followed by CaseSummarizer and SummaRuNNer. In terms of readability (Read.) as well, DSDR, CaseSummarizer and SummaRuNNer have higher mean scores than others. Finally, through the Overall ratings, we understand that DSDR is of higher satisfaction to the Law practitioners than the other algorithms, with CaseSummarizer coming second. 
These observations show a discrepancy with the automatic evaluation in Section~\ref{sec:results}  where supervised methods got better ROUGE scores than unsupervised ones.

Importantly, we again see that none of the summaries could achieve a balanced representation of all the rhetorical segments (RPC -- Arg). 
For instance, DSDR (which gets the best overall scores) represents the final judgement (RPC) and statutes (STA) well, but misses important precedents (PRE) and arguments (ARG).


In general, the experts opined that the summaries generated by several algorithms are good in the initial parts, but their quality degrades gradually from the middle. Also, the experts felt the abstractive summaries to be less organized, often having incomplete sentences; they felt that the abstractive summaries have potential but need improvement.


\vspace{2mm}
\noindent \textbf{Correlation between expert judgments and the automatic metrics:}
As stated above, there seems to be some discrepancy between expert judgements and the automatic metrics for summarization.
To explore this issue further, we compute the correlation between the expert judgments (average of the `Overall' scores of the three  annotators) and the automatic metrics (ROUGE-1,2, L Fscores and BERT-Scores). The human evaluation was conducted over 5 documents and 7 algorithms. So, for each metric, correlation was calculated between the 5 human-assigned overall scores and the 5 metric scores, and then an average was taken across all the 7 algorithms (details in Appendix Section~\ref{sec:expert-eval-details}). 

Following this procedure, the correlation of the mean `Overall' score (assigned by experts) with 
ROUGE-1 F-Score is 0.212, that with ROUGE-2 F-Score is 0.208, that with ROUGE-L F-Score is 0.132 and the correlation with BERTScore is 0.067. These low correlation scores again suggest that \textit{automatic summarization metrics may be insufficient} to judge the quality of summaries in specialized domains such as Law.

\section{Concluding discussion}


We develop datasets and benchmark results for legal case judgement summarization. 
Our study provides several guidelines for  long and legal document summarization:
(1)~For extractive summarization of legal documents, DSDR (unsupervised) and SummaRuNNer (supervised) are promising methods. 
(2)~For abstractive summarization, Legal-Pegasus (pretrained and finetuned) is a good choice.
(3)~For long documents, fine-tuning models through chunking seems a promising way. 
(4)~Document-wide evaluation does not give the complete picture; domain-specific evaluation methods, including domain experts, should also be used.

\section*{Acknowledgements}
The authors acknowledge the anonymous reviewers for their suggestions. 
The authors thank the Law domain experts from the Rajiv Gandhi School of Intellectual Property Law, India (Amritha 
Shaji, Ankita Mohanty, and Prof. Uday Shankar) and from the West Bengal National University of Juridical Sciences, India (Prof. Shouvik Guha) who helped in developing the gold standard summaries (IN-Ext dataset) and evaluating the summaries. 
The research is partially supported by the TCG Centres for Research and Education in Science and Technology (CREST) through a project titled ``Smart Legal Consultant: AI-based Legal Analytics''.

\bibliography{references}
\bibliographystyle{acl_natbib}

\clearpage

\appendix

\section{Appendix}

\subsection{Rhetorical Role Labels in a Legal Case Document}
\label{sec:rhet-roles}

According to our legal experts, rhetorical role labels/segments define a semantic function of the sentences in a legal case documents. A good summary should contain a concise representation of each segment. These rhetorical segments are defined as follows:

{\it (i)~Facts (abbreviated as FAC)}: refers to the chronology of events that led to filing the case; 

{\it (ii)~Argument (ARG)}: arguments of the contending parties; 

{\it (iii)~Statute (STA)}: Established laws referred to by the present court; 

{\it (iv)~Precedent (PRE)}: Precedents/prior cases that were referred to; 

{\it (v)~Ratio of the decision (Ratio)}: reasoning/rationale for the final judgement given by the present court; 

{\it (vi)~Ruling by Present Court (RPC)}: the final judgement given by the present court.


\subsection{Implementations details of Domain-Specific Extractive Summarization Methods} \label{sec:domain-specific-extractive-implementation}

We state here the reproducibility details of the legal domain-specific summarization methods, which could not be stated in the main paper due to lack of space.

\vspace{2mm}
\noindent $\bullet$ \textbf{Legal Dictionary}: Some domain-specific summarization methods like CaseSummarizer and Gist use a set of legal keywords for identifying importance of sentences in the input document. 
We identify these keywords using a glossary from the legal repository \url{https://www.advocatekhoj.com/library}. This website provides several legal resources for Indian legal documents, including a comprehensive glossary of legal terms.

\vspace{2mm}
\noindent $\bullet$ \textbf{MMR}: 
The original paper  experiments on BVA decision of the US jurisdiction.
The MMR method
creates a template-based summary considering various semantic parts of a legal case document, and selecting a certain number of sentences from each semantic part.
Specifically, the summary is assumed to contain 
(i)~one sentence from the procedural history, 
(ii)~one sentence from issue, (iii)~one sentence from the service history of the veteran, 
(iv)~a variable number of Reasoning \& Evidential Support sentences selected using Maximum Margin Relevance, 
(v)~one sentence from the conclusion. 
Pattern-based regex extractors are used to identify the sentences (i)-(iii) and (v). 

Reasoning \& Evidential Support sentences are identified using a 2-step supervised classification method -- in the first step, sentences predictive of a case's outcome are detected using Convolutional Neural Networks. In the second step, a Random Forest Classifier is used to specifically extract the ``Reasoning \& Evidential Support'' sentences from the predictive sentences. 
In the absence of such annotated training datasets to build a 2-stage classification framework for India and UK, we adopt only the Maximum Margin Relevance module of their work as a baseline. 

This method decides the inclusion of a sentence $S_i$ to the summary based on $\lambda \times Sim(S_i, \textit{Case}) + (1-\lambda) \times Sim(S_i, \textit{Summary})$, where \textit{Case} indicates the set of sentences in the original case document and \textit{Summary} represents the current set of sentences in the summary. $\lambda$ acts as the weight that balances the relevance and diversity; we consider $\lambda = 0.5$.

\vspace{2mm}
\noindent $\bullet$ \textbf{Gist}: Gist uses the following handcrafted features to represent every sentence in the input case document (which is to be summarized) -- \\
(i)~Quantitative features: number of words, number of characters, number of unique words, and position of the sentence \\
(ii)~Case category information: The original paper produced summaries of Chinese documents  which contain information like whether a document is recorded as a
judgment or as a ruling (which is a category of judicial judgments) and specific words that are used
by the courts to indicate subcategories of the judgments. These information are absent in Indian and UK Supreme Court Case documents. So we do not consider this category of features. \\
(iii)~Specific Legal Terms: We use a legal dictionary   for the purpose (from \url{https://www.advocatekhoj.com/library/glossary/a.php}, as stated in the main paper). \\ 
(iv)~Word Embeddings: To construct the embedding of a sentence, we take the average of the embeddings of the words in the sentence. 
To this end, we train a word2vec model on the training corpus ($7030$ documents of the IN-Abs and $693$ documents of the UK-Abs dataset). 
During evaluation, the trained word2vec model is used to derive the embeddings.\\
(v)~One-hot vectors of first $k$ POS tags in the sequence, where $k=10$ as mentioned in the paper \\ (vi)~Word Embeddings of the opening words: we take the average of the embeddings of the first 5 words in the sentence, since the paper did not clearly mention how to obtain them.

Based on the above features, Gist uses 3 models -- MLP, Gradient Boosted Decision Tree, LSTM and a combination of LSTM and MLP classifiers -- to rank sentences in order of their likelihood to be included in the summary. We observe the best performance by using Gradient Boosted Decision Tree as the ML classifier, which we report.


\vspace{2mm}
\noindent $\bullet$ \textbf{CaseSummarizer}: The original method of CaseSummarizer was developed for Australian documents. All sentences in the input document are ranked using the following score: $w_{new} = w_{old} + \sigma \left(0.2d + 0.3e + 1.5s \right)$, where 
$w_{old}$ is the sum of the TF-IDF values of its constituent words, normalized over the sentence length, 
$d$ is the number of `dates' present in the sentence, 
$e$ is the number of named entity mentions in the sentence, $s$ is a boolean variable indicating whether the sentence is at the start of any section, and 
$\sigma$ is the standard deviation among the sentence scores.  

The Indian case documents used in our study (IN-Ext and IN-Abs) are less structured than Australian case documents, and they do not contain `section headings'. 
So, in place of that feature we used a count of the number of legal terms (identified by a legal dictionary) present in the sentence. 
We could find section numbers of Acts in our gold standard summaries, for example, ``section 302 of the Indian Penal Code''. Hence, for the parameter ``d'' in the formulation, we included both dates and section numbers. 
The authors did not clearly mention how they have identified the ``entities'' in the texts. So, we have used the Stanford NER Tagger for identifying entities within the sentence. 
For ensuring a fair comparison, we have used the same setting on UK-Abs too.

\vspace{2mm}
\noindent $\bullet$ \textbf{LetSum and KMM}: Both the LetSum and KMM methods initially assign rhetorical labels to sentences (using certain cue-phrases and Conditional Random Fields respectively). The sentences are then ranked, for which LetSum uses TF-IDF scores and KMM uses a K-Mixture Model based score.  
However, the rhetorical role information is {\it not} used for generating the summary. 
Rather, the rhetorical labels are used as a {\it post-summarization step} that is mainly used for displaying the summary in a structured way. We therefore implement only the sentence ranking modules for these methods -- i.e, TF-IDF based summarization for LetSum and K-mixture model based summarization for KMM.


\subsection{Implementation Details of Domain-Independent Extractive Summarization Methods} \label{sec:general-extractive-methods-implementation}

We use the publicly available implementations of the domain-independent extractive methods from the following sources:
\begin{itemize}
    \item LexRank, LSA, Luhn and Reduction: \url{https://pypi.org/project/sumy/}
    \item PacSum: \url{https://github.com/mswellhao/PacSum}
    \item SummaRuNNer: \url{https://github.com/hpzhao/SummaRuNNer}) 
    \item BERTSUM: \url{https://github.com/nlpyang/PreSumm}. The original BERTSUM model uses a post-processing step called \textit{Trigram Blocking} that excludes a candidate sentence if it has a significant amount of trigram overlap with the already generated summary (to minimize redundancy in the summary). However, we observed that this step leads to summaries that are too short, as also observed in~\cite{ExtendedSumm}. Hence we ignore this step.
\end{itemize}


\subsection{Methods for obtaining Training Data for Extractive Supervised Methods} \label{sec:extractive-supervised-methods-training-data}

As stated in Section~\ref{sec:extractive}, we tried three methods for generating training data for extractive supervised methods from abstractive reference summaries. The best-performing \textbf{Avr} method (which we finally used in our experiments) was described in Section~\ref{sec:extractive}. Here we describe the other two methods that we tried.

\vspace{1mm}
\textbf{(i)~Maximal}: In this approach proposed in~\cite{nallapati2017summarunner} the basic premise was to maximize the ROUGE score between the extractive and the abstractive gold-standard summaries. However global optimization is computationally expensive; a faster greedy strategy is -- keep adding sentences to the extractive summary one by one, each time selecting the sentence that when added to the already extracted summary has the maximum ROUGE score with respect to the abstractive gold-standard summary. 
This process is repeated till the ROUGE score does not increase anymore. 
Finally, all the sentences in this extractive summary are labelled as 1, the rest as 0.

\vspace{1mm}
\textbf{(ii)~TF-IDF}: 
We calculated the TF-IDF vectors for all the sentences in the source document and those in the summary. 
For each sentence in the summary, we find three sentences in the full text that are most similar to it. The similarity is measured as the cosine-similarity between the TF-IDF vectors of a sentence in the summary and a sentence in the source document, and similarity should be greater than $0.4$. 
We label the sentences in the source document that are similar to some summary-sentence as 1, rest as 0.


\subsection{Implementation details of Abstractive Summarization Methods} \label{sec:abstractive-implementation}

We use the publicly available implementations of the abstractive methods from the following sources:
\begin{itemize}



\item BART: \url{https://huggingface.co/facebook/BART\_large}~\


\item Legal-Pegasus (trained on legal documents): \url{https://huggingface.co/nsi319/legal-pegasus} 

\item Legal-LED (trained on legal documents): \url{https://huggingface.co/nsi319/legal-led-base-16384}

\end{itemize}
The hyper-parameters for finetuning are given in Table~\ref{tab:hyp}.

\begin{table}[tb]
\centering
\resizebox{\linewidth}{!}{
\begin{tabular}{|c|c|}
\hline
Model  & Fine-tuning parameters                     
\\ \hline
BART    & \begin{tabular}[c]{@{}c@{}}Learning rate -  2e-5,  Epochs - 3, Batch size - 1\\ Max input length - 1024, Max output length - 512\end{tabular} \\ \hline
Legal-Pegasus & \begin{tabular}[c]{@{}c@{}}Learning rate - 5e-5, Epochs - 2, Batch size - 1\\ Max input length - 512,  Max output length - 256\end{tabular}   \\ \hline
Legal-LED & \begin{tabular}[c]{@{}c@{}}Learning rate - 1e-3, Epochs - 3, Batch size - 4\\ Max input length - 16384,  Max output length - 1024\end{tabular}   \\ \hline

\end{tabular}
}
\caption{Hyper-paramaters used in finetuning BART, Legal-Pegasus and Legal-LED.}
\label{tab:hyp}
\end{table}


\subsection{Methods for obtaining finetuning data for abstractive summarization models} \label{sec:abstractive-finetuning-methods}

As stated in Section~\ref{sec:ft-abs}, we experimented with several sentence similarity measures for generating finetuning data for abstractive models. The best performing sentence similarity measure, MCS, was described in Section~\ref{sec:ft-abs}. Here we describe the other sentence similarity measures that we tried.

\textit{(i) Smooth Inverse frequency with cosine similarity (SIF)}~\cite{ranasinghe-etal-2019-enhancing}: This approach is similar to the MCS approach; only here instead of mean, we consider a weighted mean, and we use a pre-trained BERT model. 
The weight of every token $w$ is given by $\frac{a}{a + p(w)}$
Where $p(w)$ is the estimated frequency of a word in the whole dataset.
In other word, the weight for a word would be inversely proportional to the number of word occurrences. 

\textit{(ii) Cosine similarity with BERT [CLS] token (CLS-CS):} Here we consider the cosine similarity of the encodings of the CLS tokens of the two sentences (as given by the pre-trained BERT model). 

\textit{(iii) MCS\_RR:} Here, we using Rhetorical Roles (RR) for generating finetuning data that incorporates legal domain knowledge.
As described earlier in Section~\ref{sec:dataset}, a legal case document consists of 7 rhetorical segments such as Facts, Statutes, etc.
We incorporate this knowledge into our abstractive summarization process by combining it with the divide and conquer approach presented in~\cite{divide&con} (which is originally designed for summarizing research articles that are already segmented into logical segments). 

We first use a state-of-the-art classifier for rhetorical labeling of sentences in a legal document~\cite{deeprhole} to assign one of the labels -- RPC, FAC, STA, RLC, Ratio, PRE, ARG -- to each sentence of a document. 
We collate sentences of a particular role as one segment. Thus, effectively, we partition a document into 7 segments, each segment corresponding to a rhetorical role.
Then we apply the same approach as stated above to generate the summary of each segment; for this, we use the MCS sentence similarity measure (which performs the best, as we shall see later in Section~\ref{sec:results}). 
Note that, some of these rhetorical segments themselves may be longer than  the input token limit of BART and Pegasus; in such cases, we further divide the rhetorical segments into smaller chunks, and then generate the summary of each chunk. 

\subsection{Detailed Summarization Results} \label{sec:detailed-results}

Table~\ref{tab:results-overall-in-ext-all}, Table~\ref{tab:results-overall-in-abs-all} and Table~\ref{tab:results-overall-uk-abs-all} contain the \textit{document-wide} ROUGE and BERTScores for the IN-Ext, IN-Abs and UK-Abs datasets respectively. These tables give the results for all summarization methods that we have applied (while the tables in the main text report results of only some of the best-performing methods).

Table~\ref{tab:rouge-segment-india-all} and Table~\ref{tab:rouge-segment-uk-all} contain the \textit{segment-wise} ROUGE scores over the IN-Ext and UK-Abs datasets, for all methods that we have applied.

\begin{table}[tb]
\resizebox{\linewidth}{!}{
\begin{tabular}{|c|c|c|c|c|c|}
\hline
\multicolumn{1}{|c|}{\multirow{2}{*}{\textbf{Algorithm}}} & \multicolumn{3}{c|}{\textbf{ROUGE Scores}}                                                                      & \multirow{2}{*}{\textbf{BERTScore}} \\ \cline{2-4}
\multicolumn{1}{|c|}{}                                    & \multicolumn{1}{c|}{R-1}            & \multicolumn{1}{c|}{R-2}            & \multicolumn{1}{c|}{R-L}            &                                     \\ \hline
\multicolumn{5}{|c|}{\textit{Extractive Methods}}                                                                                                                                                                 \\ \hline
\multicolumn{5}{|c|}{Unsupervised, Domain Independent}                                                                                                                                                            \\ \hline
\multicolumn{1}{|c|}{LexRank}                             & \multicolumn{1}{c|}{0.564}          & \multicolumn{1}{c|}{0.344}          & \multicolumn{1}{c|}{0.388}          & 0.862                               \\ \hline
\multicolumn{1}{|c|}{Lsa}                                 & \multicolumn{1}{c|}{0.553}          & \multicolumn{1}{c|}{0.348}          & \multicolumn{1}{c|}{0.397}          & 0.875                               \\ \hline
\multicolumn{1}{|c|}{DSDR}                                & \multicolumn{1}{c|}{0.566}          & \multicolumn{1}{c|}{0.317}          & \multicolumn{1}{c|}{0.264}          & 0.834                               \\ \hline
\multicolumn{1}{|c|}{Luhn}                                & \multicolumn{1}{c|}{0.568}          & \multicolumn{1}{c|}{0.373}          & \multicolumn{1}{c|}{\textbf{0.422}} & \textbf{0.882}                      \\ \hline
\multicolumn{1}{|c|}{Reduction}                           & \multicolumn{1}{c|}{0.561}          & \multicolumn{1}{c|}{0.358}          & \multicolumn{1}{c|}{0.405}          & 0.869                               \\ \hline
\multicolumn{1}{|c|}{Pacsum\_bert}                        & \multicolumn{1}{c|}{\textbf{0.590}}  & \multicolumn{1}{c|}{\textbf{0.410}}  & \multicolumn{1}{c|}{0.335}          & 0.879                               \\ \hline
\multicolumn{1}{|c|}{Pacsum\_tfidf}                       & \multicolumn{1}{c|}{0.566}          & \multicolumn{1}{c|}{0.357}          & \multicolumn{1}{c|}{0.301}          & 0.839                               \\ \hline
\multicolumn{5}{|c|}{Unsupervised, Legal Domain Specific}                                                                                                                                                         \\ \hline
\multicolumn{1}{|c|}{MMR}                                 & \multicolumn{1}{c|}{0.563}          & \multicolumn{1}{c|}{0.318}          & \multicolumn{1}{c|}{0.262}          & 0.833                               \\ \hline
\multicolumn{1}{|c|}{KMM}                      & \multicolumn{1}{c|}{0.532}          & \multicolumn{1}{c|}{0.302}          & \multicolumn{1}{c|}{0.28}           & 0.836                               \\ \hline
\multicolumn{1}{|c|}{LetSum}                              & \multicolumn{1}{c|}{\textbf{0.591}} & \multicolumn{1}{c|}{\textbf{0.401}} & \multicolumn{1}{c|}{\textbf{0.391}} & \textbf{0.875}                      \\ \hline
\multicolumn{1}{|c|}{CaseSummarizer}                      & \multicolumn{1}{c|}{0.52}           & \multicolumn{1}{c|}{0.321}          & \multicolumn{1}{c|}{0.279}          & 0.835                               \\ \hline
\multicolumn{5}{|c|}{Supervised, Domain Independent}                                                                                                                                                              \\ \hline
\multicolumn{1}{|c|}{SummaRunner}                         & \multicolumn{1}{c|}{0.532}          & \multicolumn{1}{c|}{0.334}          & \multicolumn{1}{c|}{0.269}          & 0.829                               \\ \hline
\multicolumn{1}{|c|}{BERT-Ext}                            & \multicolumn{1}{c|}{\textbf{0.589}} & \multicolumn{1}{c|}{\textbf{0.398}} & \multicolumn{1}{c|}{\textbf{0.292}} & \textbf{0.85}                       \\ \hline
\multicolumn{5}{|c|}{Supervised, Legal Domain Specific}                                                                                                                                                           \\ \hline
\multicolumn{1}{|c|}{Gist}                                & \multicolumn{1}{c|}{0.555}          & \multicolumn{1}{c|}{0.335}          & \multicolumn{1}{c|}{0.391}          & 0.864                               \\ \hline
\multicolumn{5}{|c|}{\textit{Abstractive Methods}}                                                                                                                                                                \\ \hline
\multicolumn{5}{|c|}{Pretrained}                                                                                                                                                                                  \\ \hline
\multicolumn{1}{|c|}{BART}                                & \multicolumn{1}{c|}{0.475}          & \multicolumn{1}{c|}{0.221}          & \multicolumn{1}{c|}{0.271}          & 0.833                               \\ \hline
\multicolumn{1}{|c|}{BERT-BART}                           & \multicolumn{1}{c|}{\textbf{0.488}} & \multicolumn{1}{c|}{\textbf{0.236}} & \multicolumn{1}{c|}{\textbf{0.279}} & 0.836                               \\ \hline
\multicolumn{1}{|c|}{Legal-Pegasus}                             & \multicolumn{1}{c|}{0.465}          & \multicolumn{1}{c|}{0.211}          & \multicolumn{1}{c|}{\textbf{0.279}} & \textbf{0.842}                      \\ \hline
\multicolumn{1}{|c|}{Legal-LED}                          & \multicolumn{1}{c|}{0.175}          & \multicolumn{1}{c|}{0.036}          & \multicolumn{1}{c|}{0.12}           & 0.799                               \\ \hline
\multicolumn{5}{|c|}{Finetuned}                                                                                                                                                                    \\ \hline
\multicolumn{1}{|c|}{BART\_CLS}                           & \multicolumn{1}{c|}{0.534}          & \multicolumn{1}{c|}{0.29}           & \multicolumn{1}{c|}{0.349}          & 0.853                               \\ \hline
\multicolumn{1}{|c|}{BART\_MCS}                           & \multicolumn{1}{c|}{0.557}          & \multicolumn{1}{c|}{0.322}          & \multicolumn{1}{c|}{0.404}          & 0.868                               \\ \hline
\multicolumn{1}{|c|}{BART\_SIF}                           & \multicolumn{1}{c|}{0.540}           & \multicolumn{1}{c|}{0.304}          & \multicolumn{1}{c|}{0.369}          & 0.857                               \\ \hline
\multicolumn{1}{|c|}{BERT\_BART\_MCS}                     & \multicolumn{1}{c|}{0.553}          & \multicolumn{1}{c|}{0.316}          & \multicolumn{1}{c|}{0.403}          & \textbf{0.869}                      \\ \hline
\multicolumn{1}{|c|}{Legal-Pegasus\_MCS}                        & \multicolumn{1}{c|}{\textbf{0.575}} & \multicolumn{1}{c|}{\textbf{0.351}} & \multicolumn{1}{c|}{\textbf{0.419}} & 0.864                               \\ \hline
\multicolumn{1}{|c|}{Legal-LED}                          & \multicolumn{1}{c|}{0.471}          & \multicolumn{1}{c|}{0.26}           & \multicolumn{1}{c|}{0.341}          & 0.863                               \\ \hline
\multicolumn{1}{|c|}{BART\_MCS\_RR}                       & \multicolumn{1}{c|}{0.574}          & \multicolumn{1}{c|}{0.345}          & \multicolumn{1}{c|}{0.402}          & 0.864                               \\ \hline
\end{tabular}
}
\caption{Document-wide ROUGE-L and BERTScores (Fscore) on the IN-Ext dataset. All values averaged over the $50$ documents in the dataset. The best value in a particular class of methods is in \textbf{bold}.}
\label{tab:results-overall-in-ext-all}
\vspace{3mm}
\end{table}

\begin{table}[tb]
\resizebox{\linewidth}{!}{
\begin{tabular}{|ccccc|}
\hline
\multicolumn{1}{|c|}{\multirow{2}{*}{\textbf{Algorithm}}} & \multicolumn{3}{c|}{\textbf{ROUGE Scores}}                                                                      & \multirow{2}{*}{\textbf{BERTScore}} \\ \cline{2-4}
\multicolumn{1}{|c|}{}                                    & \multicolumn{1}{c|}{R-1}            & \multicolumn{1}{c|}{R-2}            & \multicolumn{1}{c|}{R-L}            &                                     \\ \hline
\multicolumn{5}{|c|}{\textit{Extractive Methods}}                                                                                                                                                                 \\ \hline
\multicolumn{5}{|c|}{Unsupervised, Domain Independent}                                                                                                                                                            \\ \hline
\multicolumn{1}{|c|}{LexRank}                             & \multicolumn{1}{c|}{0.436}          & \multicolumn{1}{c|}{0.195}          & \multicolumn{1}{c|}{\textbf{0.284}} & 0.843                               \\ \hline
\multicolumn{1}{|c|}{Lsa}                                 & \multicolumn{1}{c|}{0.401}          & \multicolumn{1}{c|}{0.172}          & \multicolumn{1}{c|}{0.259}          & 0.834                               \\ \hline
\multicolumn{1}{|c|}{DSDR}                                & \multicolumn{1}{c|}{\textbf{0.485}} & \multicolumn{1}{c|}{\textbf{0.222}} & \multicolumn{1}{c|}{0.27}           & \textbf{0.848}                      \\ \hline
\multicolumn{1}{|c|}{Luhn}                                & \multicolumn{1}{c|}{0.405}          & \multicolumn{1}{c|}{0.181}          & \multicolumn{1}{c|}{0.268}          & 0.837                               \\ \hline
\multicolumn{1}{|c|}{Reduction}                           & \multicolumn{1}{c|}{0.431}          & \multicolumn{1}{c|}{0.195}          & \multicolumn{1}{c|}{\textbf{0.284}} & 0.844                               \\ \hline
\multicolumn{1}{|c|}{Pacsum\_bert}                        & \multicolumn{1}{c|}{0.401}          & \multicolumn{1}{c|}{0.175}          & \multicolumn{1}{c|}{0.242}          & 0.839                               \\ \hline
\multicolumn{1}{|c|}{Pacsum\_tfidf}                       & \multicolumn{1}{c|}{0.428}          & \multicolumn{1}{c|}{0.194}          & \multicolumn{1}{c|}{0.262}          & 0.834                               \\ \hline
\multicolumn{5}{|c|}{Unsupervised, Legal Domain Specific}                                                                                                                                                         \\ \hline
\multicolumn{1}{|c|}{MMR}                                 & \multicolumn{1}{c|}{0.452}          & \multicolumn{1}{c|}{0.21}           & \multicolumn{1}{c|}{0.253}          & \textbf{0.844}                      \\ \hline
\multicolumn{1}{|c|}{KMM}                      & \multicolumn{1}{c|}{\textbf{0.455}} & \multicolumn{1}{c|}{0.2}            & \multicolumn{1}{c|}{0.259}          & 0.843                               \\ \hline
\multicolumn{1}{|c|}{LetSum}                              & \multicolumn{1}{c|}{0.395}          & \multicolumn{1}{c|}{0.167}          & \multicolumn{1}{c|}{0.251}          & 0.833                               \\ \hline
\multicolumn{1}{|c|}{CaseSummarizer}                      & \multicolumn{1}{c|}{0.454}          & \multicolumn{1}{c|}{\textbf{0.229}} & \multicolumn{1}{c|}{\textbf{0.279}} & 0.843                               \\ \hline
\multicolumn{5}{|c|}{Supervised, Domain Independent}                                                                                                                                                              \\ \hline
\multicolumn{1}{|c|}{SummaRunner}                         & \multicolumn{1}{c|}{\textbf{0.493}} & \multicolumn{1}{c|}{\textbf{0.255}} & \multicolumn{1}{c|}{\textbf{0.274}} & \textbf{0.849}                      \\ \hline
\multicolumn{1}{|c|}{BERT-Ext}                            & \multicolumn{1}{c|}{0.427}          & \multicolumn{1}{c|}{0.199}          & \multicolumn{1}{c|}{0.239}          & 0.821                               \\ \hline
\multicolumn{5}{|c|}{Supervised, Legal Domain Specific}                                                                                                                                                           \\ \hline
\multicolumn{1}{|c|}{Gist}                                & \multicolumn{1}{c|}{0.471}          & \multicolumn{1}{c|}{0.238}          & \multicolumn{1}{c|}{0.308}          & 0.842                               \\ \hline
\multicolumn{5}{|c|}{\textit{Abstractive Methods}}                                                                                                                                                                \\ \hline
\multicolumn{5}{|c|}{Pretrained}                                                                                                                                                                                  \\ \hline
\multicolumn{1}{|c|}{BART}                                & \multicolumn{1}{c|}{0.39}           & \multicolumn{1}{c|}{0.156}          & \multicolumn{1}{c|}{0.246}          & 0.829                               \\ \hline
\multicolumn{1}{|c|}{BERT-BART}                           & \multicolumn{1}{c|}{0.337}          & \multicolumn{1}{c|}{0.112}          & \multicolumn{1}{c|}{0.212}          & 0.809                               \\ \hline
\multicolumn{1}{|c|}{Legal-Pegasus}                             & \multicolumn{1}{c|}{\textbf{0.441}} & \multicolumn{1}{c|}{\textbf{0.19}}  & \multicolumn{1}{c|}{\textbf{0.278}} & \textbf{0.845}                      \\ \hline
\multicolumn{1}{|c|}{Legal-LED}                          & \multicolumn{1}{c|}{0.223}          & \multicolumn{1}{c|}{0.053}          & \multicolumn{1}{c|}{0.159}          & 0.813                               \\ \hline
\multicolumn{5}{|c|}{Finetuned}                                                                                                                                                                                   \\ \hline
\multicolumn{1}{|c|}{BART\_CLS}                           & \multicolumn{1}{c|}{0.484}          & \multicolumn{1}{c|}{0.231}          & \multicolumn{1}{c|}{0.311}          & 0.85                                \\ \hline
\multicolumn{1}{|c|}{BART\_MCS}                           & \multicolumn{1}{c|}{\textbf{0.495}} & \multicolumn{1}{c|}{0.249}          & \multicolumn{1}{c|}{0.33}           & 0.851                               \\ \hline
\multicolumn{1}{|c|}{BART\_SIF}                           & \multicolumn{1}{c|}{0.49}           & \multicolumn{1}{c|}{0.246}          & \multicolumn{1}{c|}{0.326}          & 0.851                               \\ \hline
\multicolumn{1}{|c|}{BERT\_BART\_MCS}                     & \multicolumn{1}{c|}{0.487}          & \multicolumn{1}{c|}{0.243}          & \multicolumn{1}{c|}{0.329}          & 0.853                               \\ \hline
\multicolumn{1}{|c|}{Legal-Pegasus\_MCS}                         & \multicolumn{1}{c|}{0.488}          & \multicolumn{1}{c|}{\textbf{0.252}} & \multicolumn{1}{c|}{\textbf{0.341}} & 0.851                               \\ \hline
\multicolumn{1}{|c|}{Legal-LED}                          & \multicolumn{1}{c|}{0.471}          & \multicolumn{1}{c|}{0.235}          & \multicolumn{1}{c|}{0.332}          & \textbf{0.856}                      \\ \hline
\multicolumn{1}{|c|}{BART\_MCS\_RR}                       & \multicolumn{1}{c|}{0.49}           & \multicolumn{1}{c|}{0.234}          & \multicolumn{1}{c|}{0.311}          & 0.849                               \\ \hline
\end{tabular}
}
\caption{Document-wide ROUGE-L and BERTScores (Fscore) on the IN-Abs dataset, averaged over the $100$ test documents. 
The best value in a particular class of methods is in \textbf{bold}.
}
\label{tab:results-overall-in-abs-all}
\vspace{3mm}
\end{table}

\begin{table}[tb]
\resizebox{\linewidth}{!}{
\begin{tabular}{|ccccc|}
\hline
\multicolumn{1}{|c|}{\multirow{2}{*}{\textbf{Algorithm}}} & \multicolumn{3}{c|}{\textbf{ROUGE Scores}}                                                                      & \multirow{2}{*}{\textbf{BERTScore}} \\ \cline{2-4}
\multicolumn{1}{|c|}{}                                    & \multicolumn{1}{c|}{R-1}            & \multicolumn{1}{c|}{R-2}            & \multicolumn{1}{c|}{R-L}            &                                     \\ \hline
\multicolumn{5}{|c|}{\textit{Extractive Methods}}                                                                                                                                                                 \\ \hline
\multicolumn{5}{|c|}{Unsupervised, Domain Independent}                                                                                                                                                            \\ \hline
\multicolumn{1}{|c|}{LexRank}                             & \multicolumn{1}{c|}{0.481}          & \multicolumn{1}{c|}{\textbf{0.187}} & \multicolumn{1}{c|}{\textbf{0.265}} & \textbf{0.848}                      \\ \hline
\multicolumn{1}{|c|}{Lsa}                                 & \multicolumn{1}{c|}{0.426}          & \multicolumn{1}{c|}{0.149}          & \multicolumn{1}{c|}{0.236}          & 0.843                               \\ \hline
\multicolumn{1}{|c|}{DSDR}                                & \multicolumn{1}{c|}{\textbf{0.484}} & \multicolumn{1}{c|}{0.174}          & \multicolumn{1}{c|}{0.221}          & 0.832                               \\ \hline
\multicolumn{1}{|c|}{Luhn}                                & \multicolumn{1}{c|}{0.444}          & \multicolumn{1}{c|}{0.171}          & \multicolumn{1}{c|}{0.25}           & 0.844                               \\ \hline
\multicolumn{1}{|c|}{Reduction}                           & \multicolumn{1}{c|}{0.447}          & \multicolumn{1}{c|}{0.169}          & \multicolumn{1}{c|}{0.253}          & 0.844                               \\ \hline
\multicolumn{1}{|c|}{Pacsum\_bert}                        & \multicolumn{1}{c|}{0.448}          & \multicolumn{1}{c|}{0.175}          & \multicolumn{1}{c|}{0.228}          & 0.843                               \\ \hline
\multicolumn{1}{|c|}{Pacsum\_tfidf}                       & \multicolumn{1}{c|}{0.414}          & \multicolumn{1}{c|}{0.146}          & \multicolumn{1}{c|}{0.213}          & 0.825                               \\ \hline
\multicolumn{5}{|c|}{Unsupervised, Legal Domain Specific}                                                                                                                                                         \\ \hline
\multicolumn{1}{|c|}{MMR}                                 & \multicolumn{1}{c|}{0.440}           & \multicolumn{1}{c|}{0.151}          & \multicolumn{1}{c|}{0.205}          & 0.83                                \\ \hline
\multicolumn{1}{|c|}{KMM}                      & \multicolumn{1}{c|}{0.430}           & \multicolumn{1}{c|}{0.138}          & \multicolumn{1}{c|}{0.201}          & 0.827                               \\ \hline
\multicolumn{1}{|c|}{LetSum}                              & \multicolumn{1}{c|}{0.437}          & \multicolumn{1}{c|}{0.158}          & \multicolumn{1}{c|}{\textbf{0.233}} & \textbf{0.842}                      \\ \hline
\multicolumn{1}{|c|}{CaseSummarizer}                      & \multicolumn{1}{c|}{\textbf{0.445}} & \multicolumn{1}{c|}{\textbf{0.166}} & \multicolumn{1}{c|}{0.227}          & 0.835                               \\ \hline
\multicolumn{5}{|c|}{Supervised, Domain Independent}                                                                                                                                                              \\ \hline
\multicolumn{1}{|c|}{SummaRunner}                         & \multicolumn{1}{c|}{\textbf{0.502}} & \multicolumn{1}{c|}{\textbf{0.205}} & \multicolumn{1}{c|}{\textbf{0.237}} & \textbf{0.846}                      \\ \hline
\multicolumn{1}{|c|}{BERT-Ext}                            & \multicolumn{1}{c|}{0.431}          & \multicolumn{1}{c|}{0.184}          & \multicolumn{1}{c|}{0.24}           & 0.821                               \\ \hline
\multicolumn{5}{|c|}{Supervised, Legal Domain Specific}                                                                                                                                                           \\ \hline
\multicolumn{1}{|c|}{Gist}                                & \multicolumn{1}{c|}{0.427}          & \multicolumn{1}{c|}{0.132}          & \multicolumn{1}{c|}{0.215}          & 0.819                               \\ \hline
\multicolumn{5}{|c|}{\textit{Abstractive Methods}}                                                                                                                                                                \\ \hline
\multicolumn{5}{|c|}{Pretrained}                                                                                                                                                                                  \\ \hline
\multicolumn{1}{|c|}{Pointer\_Generator}                  & \multicolumn{1}{c|}{0.420}           & \multicolumn{1}{c|}{0.133}          & \multicolumn{1}{c|}{0.193}          & 0.812                               \\ \hline
\multicolumn{1}{|c|}{BERT-Abs}                            & \multicolumn{1}{c|}{0.362}          & \multicolumn{1}{c|}{0.087}          & \multicolumn{1}{c|}{0.208}          & 0.803                               \\ \hline
\multicolumn{1}{|c|}{BART}                                & \multicolumn{1}{c|}{0.436}          & \multicolumn{1}{c|}{0.142}          & \multicolumn{1}{c|}{0.236}          & 0.837                               \\ \hline
\multicolumn{1}{|c|}{BERT-BART}                           & \multicolumn{1}{c|}{0.369}          & \multicolumn{1}{c|}{0.099}          & \multicolumn{1}{c|}{0.198}          & 0.805                               \\ \hline
\multicolumn{1}{|c|}{Legal-Pegasus}                             & \multicolumn{1}{c|}{\textbf{0.452}} & \multicolumn{1}{c|}{\textbf{0.155}} & \multicolumn{1}{c|}{\textbf{0.248}} & \textbf{0.843}                      \\ \hline
\multicolumn{1}{|c|}{Legal-LED}                          & \multicolumn{1}{c|}{0.197}          & \multicolumn{1}{c|}{0.038}          & \multicolumn{1}{c|}{0.138}          & 0.814                               \\ \hline
\multicolumn{5}{|c|}{Finetuned}                              \\ \hline
\multicolumn{1}{|c|}{BART\_CLS}                           & \multicolumn{1}{c|}{0.481}          & \multicolumn{1}{c|}{0.172}          & \multicolumn{1}{c|}{0.255}          & 0.844                               \\ \hline
\multicolumn{1}{|c|}{BART\_MCS}                           & \multicolumn{1}{c|}{\textbf{0.496}} & \multicolumn{1}{c|}{\textbf{0.188}} & \multicolumn{1}{c|}{\textbf{0.271}}          & \textbf{0.848}                      \\ \hline
\multicolumn{1}{|c|}{BART\_SIF}                           & \multicolumn{1}{c|}{0.485}          & \multicolumn{1}{c|}{0.18}           & \multicolumn{1}{c|}{0.262}          & 0.845                               \\ \hline
\multicolumn{1}{|c|}{BERT\_BART\_MCS}                     & \multicolumn{1}{c|}{0.476}          & \multicolumn{1}{c|}{0.172}          & \multicolumn{1}{c|}{0.259}          & 0.847                               \\ \hline
\multicolumn{1}{|c|}{Legal-Pegasus\_MCS}                         & \multicolumn{1}{c|}{0.476}          & \multicolumn{1}{c|}{0.171}          & \multicolumn{1}{c|}{0.261}          & 0.838                               \\ \hline
\multicolumn{1}{|c|}{Legal-LED}                          & \multicolumn{1}{c|}{0.482}          & \multicolumn{1}{c|}{0.186}          & \multicolumn{1}{c|}{0.264}          & 0.851                               \\ \hline
\multicolumn{1}{|c|}{BART\_MCS\_RR}                       & \multicolumn{1}{c|}{0.492}          & \multicolumn{1}{c|}{0.184}          & \multicolumn{1}{c|}{0.26}           & 0.839                               \\ \hline
\end{tabular}
}
\caption{Document-wide ROUGE-L and BERTScores (Fscore) on UK-Abs dataset, averaged over the $100$ test documents. The best value for each category of methods is in \textbf{bold}.}
\label{tab:results-overall-uk-abs-all}
\vspace{4mm}
\end{table}

\begin{table}[tb]
\resizebox{\linewidth}{!}{
\begin{tabular}{|cccccc|}
\hline
\multicolumn{1}{|c|}{\multirow{2}{*}{\textbf{Algorithms}}} &  \multicolumn{5}{c|}{\textbf{Rouge L Recall}} \\ \cline{2-6} 
\multicolumn{1}{|c|}{} & \multicolumn{1}{c|}{\begin{tabular}[c]{@{}c@{}}\textbf{RPC}\\ (6.42\%)\end{tabular}} &  \multicolumn{1}{c|}{\begin{tabular}[c]{@{}c@{}}\textbf{FAC}\\ (34.85\%)\end{tabular}} &  \multicolumn{1}{c|}{\begin{tabular}[c]{@{}c@{}}\textbf{STA}\\ (13.42\%)\end{tabular}} & \multicolumn{1}{c|}{\begin{tabular}[c]{@{}c@{}}\textbf{Ratio+Pre}\\ (28.83\%)\end{tabular}} &  \multicolumn{1}{c|}{\begin{tabular}[c]{@{}c@{}}\textbf{ARG}\\ (16.45\%)\end{tabular}} \\ \hline
\multicolumn{6}{|c|}{\textit{Extractive Methods}}                                                                                                                                                                                  \\ \hline
\multicolumn{1}{|c|}{LexRank}                             & \multicolumn{1}{c|}{0.039}          & \multicolumn{1}{c|}{0.204}          & \multicolumn{1}{c|}{0.104}          & \multicolumn{1}{c|}{0.208}          & \textbf{0.127} \\ \hline
\multicolumn{1}{|c|}{Lsa}                                 & \multicolumn{1}{c|}{0.037}          & \multicolumn{1}{c|}{0.241}          & \multicolumn{1}{c|}{0.091}          & \multicolumn{1}{c|}{0.188}          & 0.114          \\ \hline
\multicolumn{1}{|c|}{DSDR}                                & \multicolumn{1}{c|}{0.053}          & \multicolumn{1}{c|}{0.144}          & \multicolumn{1}{c|}{0.099}          & \multicolumn{1}{c|}{0.21}           & 0.104          \\ \hline
\multicolumn{1}{|c|}{Luhn}                                & \multicolumn{1}{c|}{0.037}          & \multicolumn{1}{c|}{\textbf{0.272}} & \multicolumn{1}{c|}{0.097}          & \multicolumn{1}{c|}{0.175}          & 0.117          \\ \hline
\multicolumn{1}{|c|}{Reduction}                           & \multicolumn{1}{c|}{0.038}          & \multicolumn{1}{c|}{0.236}          & \multicolumn{1}{c|}{0.101}          & \multicolumn{1}{c|}{0.196}          & 0.119          \\ \hline
\multicolumn{1}{|c|}{Pacsum\_bert}                        & \multicolumn{1}{c|}{0.038}          & \multicolumn{1}{c|}{0.238}          & \multicolumn{1}{c|}{0.087}          & \multicolumn{1}{c|}{0.154}          & 0.113          \\ \hline
\multicolumn{1}{|c|}{Pacsum\_tfidf}                       & \multicolumn{1}{c|}{0.039}          & \multicolumn{1}{c|}{0.189}          & \multicolumn{1}{c|}{0.111}          & \multicolumn{1}{c|}{0.18}           & 0.111          \\ \hline
\multicolumn{1}{|c|}{MMR}                                 & \multicolumn{1}{c|}{0.049}          & \multicolumn{1}{c|}{0.143}          & \multicolumn{1}{c|}{0.092}          & \multicolumn{1}{c|}{0.198}          & 0.096          \\ \hline
\multicolumn{1}{|c|}{KMM}                                 & \multicolumn{1}{c|}{0.049}          & \multicolumn{1}{c|}{0.143}          & \multicolumn{1}{c|}{0.1}            & \multicolumn{1}{c|}{0.198}          & 0.103          \\ \hline
\multicolumn{1}{|c|}{LetSum}                              & \multicolumn{1}{c|}{0.036}          & \multicolumn{1}{c|}{0.237}          & \multicolumn{1}{c|}{\textbf{0.115}} & \multicolumn{1}{c|}{0.189}          & 0.1            \\ \hline
\multicolumn{1}{|c|}{CaseSummarizer}                      & \multicolumn{1}{c|}{0.044}          & \multicolumn{1}{c|}{0.148}          & \multicolumn{1}{c|}{0.084}          & \multicolumn{1}{c|}{0.212}          & 0.104          \\ \hline
\multicolumn{1}{|c|}{SummaRunner}                         & \multicolumn{1}{c|}{\textbf{0.059}} & \multicolumn{1}{c|}{0.158}          & \multicolumn{1}{c|}{0.08}           & \multicolumn{1}{c|}{0.209}          & 0.096          \\ \hline
\multicolumn{1}{|c|}{BERT-Ext}                            & \multicolumn{1}{c|}{0.038}          & \multicolumn{1}{c|}{0.199}          & \multicolumn{1}{c|}{0.082}          & \multicolumn{1}{c|}{0.162}          & 0.093          \\ \hline
\multicolumn{1}{|c|}{Gist}                                & \multicolumn{1}{c|}{0.041}          & \multicolumn{1}{c|}{0.191}          & \multicolumn{1}{c|}{0.102}          & \multicolumn{1}{c|}{\textbf{0.223}} & 0.093          \\ \hline
\multicolumn{6}{|c|}{\textit{Pretrained Abstractive Methods}}                                                                                                                                                                      \\ \hline
\multicolumn{1}{|c|}{BART}                                & \multicolumn{1}{c|}{0.037}          & \multicolumn{1}{c|}{0.148}          & \multicolumn{1}{c|}{0.076}          & \multicolumn{1}{c|}{0.187}          & 0.087          \\ \hline
\multicolumn{1}{|c|}{BERT-BART}                           & \multicolumn{1}{c|}{0.038}          & \multicolumn{1}{c|}{0.154}          & \multicolumn{1}{c|}{0.078}          & \multicolumn{1}{c|}{0.187}          & 0.084          \\ \hline
\multicolumn{1}{|c|}{Legal-Pegasus}                             & \multicolumn{1}{c|}{0.043}          & \multicolumn{1}{c|}{0.139}          & \multicolumn{1}{c|}{0.076}          & \multicolumn{1}{c|}{0.186}          & 0.092          \\ \hline
\multicolumn{1}{|c|}{Legal-LED}                          & \multicolumn{1}{c|}{0.049}          & \multicolumn{1}{c|}{0.131}          & \multicolumn{1}{c|}{0.078}          & \multicolumn{1}{c|}{0.228}          & 0.091          \\ \hline
\multicolumn{6}{|c|}{\textit{Finetuned Abstractive Methods}}                                                                                                                                                                       \\ \hline
\multicolumn{1}{|c|}{BART\_MCS}                           & \multicolumn{1}{c|}{0.036}          & \multicolumn{1}{c|}{0.206}          & \multicolumn{1}{c|}{0.082}          & \multicolumn{1}{c|}{0.228}          & 0.092          \\ \hline
\multicolumn{1}{|c|}{BERT\_BART\_MCS}                     & \multicolumn{1}{c|}{0.037}          & \multicolumn{1}{c|}{0.205}          & \multicolumn{1}{c|}{0.085}          & \multicolumn{1}{c|}{0.237}          & 0.094          \\ \hline
\multicolumn{1}{|c|}{Legal-Pegasus\_MCS}                        & \multicolumn{1}{c|}{0.037}          & \multicolumn{1}{c|}{0.192}          & \multicolumn{1}{c|}{\textbf{0.09}}  & \multicolumn{1}{c|}{\textbf{0.257}} & 0.101 \\ \hline
\multicolumn{1}{|c|}{Legal-LED}                          & \multicolumn{1}{c|}{0.053}          & \multicolumn{1}{c|}{\textbf{0.245}} & \multicolumn{1}{c|}{0.086}          & \multicolumn{1}{c|}{0.187}          & \textbf{0.124}          \\ \hline
\multicolumn{1}{|c|}{BART\_MCS\_RR}                       & \multicolumn{1}{c|}{\textbf{0.061}} & \multicolumn{1}{c|}{0.192}          & \multicolumn{1}{c|}{0.082}          & \multicolumn{1}{c|}{0.237}          & 0.086          \\ \hline
\end{tabular}
}
\caption{Segment-wise ROUGE-L Recall scores of all methods on the IN-Ext dataset. All values averaged over the $50$ documents in the dataset. The best value for each segment in a particular class of methods is in \textbf{bold}.}
\label{tab:rouge-segment-india-all}
\vspace{4mm}
\end{table}

\begin{table}[tb]
\resizebox{\linewidth}{!}{
\begin{tabular}{|c|c|c|c|}
\hline
\multicolumn{1}{|c|}{\multirow{2}{*}{\textbf{Algorithms}}}  & \multicolumn{3}{c|}{\textbf{Rouge-L Recall}} \\ \cline{2-4}
 & \multicolumn{1}{c|}{\begin{tabular}[c]{@{}c@{}}\textbf{Background}\\ (39\%)\end{tabular}} & \multicolumn{1}{c|}{\begin{tabular}[c]{@{}c@{}}\textbf{Final Judgement}\\ (5\%)\end{tabular}} & \begin{tabular}[c]{@{}c@{}}\textbf{Reasons}\\ (56\%)\end{tabular} \\ \hline

\multicolumn{4}{|c|}{\textit{Extractive Methods}}                                                                                                       \\ \hline
\multicolumn{1}{|c|}{LexRank}                              & \multicolumn{1}{c|}{0.197}          & \multicolumn{1}{c|}{0.037}          & 0.161          \\ \hline
\multicolumn{1}{|c|}{Lsa}                                  & \multicolumn{1}{c|}{0.175}          & \multicolumn{1}{c|}{0.036}          & 0.141          \\ \hline
\multicolumn{1}{|c|}{DSDR}                                 & \multicolumn{1}{c|}{0.151}          & \multicolumn{1}{c|}{0.041}          & 0.178          \\ \hline
\multicolumn{1}{|c|}{Luhn}                                 & \multicolumn{1}{c|}{0.193}          & \multicolumn{1}{c|}{0.034}          & 0.146          \\ \hline
\multicolumn{1}{|c|}{Reduction}                            & \multicolumn{1}{c|}{0.188}          & \multicolumn{1}{c|}{0.035}          & 0.158          \\ \hline
\multicolumn{1}{|c|}{Pacsum\_bert}                         & \multicolumn{1}{c|}{0.176}          & \multicolumn{1}{c|}{0.036}          & 0.148          \\ \hline
\multicolumn{1}{|c|}{Pacsum\_tfidf}                        & \multicolumn{1}{c|}{0.154}          & \multicolumn{1}{c|}{0.035}          & 0.157          \\ \hline
\multicolumn{1}{|c|}{MMR}                                  & \multicolumn{1}{c|}{0.152}          & \multicolumn{1}{c|}{0.04}           & 0.17           \\ \hline
\multicolumn{1}{|c|}{KMM}                                  & \multicolumn{1}{c|}{0.133}          & \multicolumn{1}{c|}{0.037}          & 0.157          \\ \hline
\multicolumn{1}{|c|}{LetSum}                               & \multicolumn{1}{c|}{0.133}          & \multicolumn{1}{c|}{0.037}          & 0.147          \\ \hline
\multicolumn{1}{|c|}{CaseSummarizer}                       & \multicolumn{1}{c|}{0.153}          & \multicolumn{1}{c|}{0.036}          & 0.17           \\ \hline
\multicolumn{1}{|c|}{SummaRunner}                          & \multicolumn{1}{c|}{0.172}          & \multicolumn{1}{c|}{\textbf{0.044}} & 0.165          \\ \hline
\multicolumn{1}{|c|}{BERT-Ext}                             & \multicolumn{1}{c|}{\textbf{0.203}} & \multicolumn{1}{c|}{0.034}          & 0.135          \\ \hline
\multicolumn{1}{|c|}{Gist}                                 & \multicolumn{1}{c|}{0.123}          & \multicolumn{1}{c|}{0.041}          & \textbf{0.195} \\ \hline
\multicolumn{4}{|c|}{\textit{Pretrained Abstractive Methods}}                                                                                           \\ \hline
\multicolumn{1}{|c|}{BART}                                 & \multicolumn{1}{c|}{0.161}          & \multicolumn{1}{c|}{0.04}           & 0.175          \\ \hline
\multicolumn{1}{|c|}{BERT-BART}                            & \multicolumn{1}{c|}{0.143}          & \multicolumn{1}{c|}{0.04}           & 0.158          \\ \hline
\multicolumn{1}{|c|}{Legal-Pegasus}                              & \multicolumn{1}{c|}{0.169}          & \multicolumn{1}{c|}{0.042}          & 0.177          \\ \hline
\multicolumn{1}{|c|}{Legal-LED}                           & \multicolumn{1}{c|}{\textbf{0.177}} & \multicolumn{1}{c|}{\textbf{0.066}} & \textbf{0.219} \\ \hline
\multicolumn{4}{|c|}{\textit{Finetuned Abstractive Methods}}                                                                                            \\ \hline
\multicolumn{1}{|c|}{BART\_MCS}                                 & \multicolumn{1}{c|}{0.168}          & \multicolumn{1}{c|}{0.041}          & 0.184          \\ \hline
\multicolumn{1}{|c|}{BERT\_BART\_MCS}                           & \multicolumn{1}{c|}{0.174}          & \multicolumn{1}{c|}{0.047}          & 0.183          \\ \hline
\multicolumn{1}{|c|}{Legal-Pegasus\_MCS}                              & \multicolumn{1}{c|}{0.166}          & \multicolumn{1}{c|}{0.039}          & \textbf{0.202} \\ \hline
\multicolumn{1}{|c|}{Legal-LED}                           & \multicolumn{1}{c|}{\textbf{0.187}} & \multicolumn{1}{c|}{\textbf{0.058}} & 0.172          \\ \hline
\multicolumn{1}{|c|}{BART\_MCS\_RR}                             & \multicolumn{1}{c|}{0.165}          & \multicolumn{1}{c|}{0.042}          & 0.18           \\ \hline
\end{tabular}
}
\caption{Segment-wise ROUGE-L Recall scores of all methods on the UK-Abs dataset. All values averaged over $100$ documents in the evaluation set. Best value for each segment in a particular class of methods is in \textbf{bold}.}
\label{tab:rouge-segment-uk-all}
\vspace{4mm}
\end{table}


\subsection{More Insights from Segment-wise Evaluation} \label{sec:segment-wise-eval-detailed-insights}

Table~\ref{tab:rouge-segment-india-all} shows the segment-wise ROUGE-L Recall scores of all methods on the IN-Ext dataset, considering the 5 rhetorical segments RPC, FAC, STA, ARG, and Ratio+PRE. 
Similarly, Table~\ref{tab:rouge-segment-uk-all} shows the segment-wise ROUGE-L Recall scores of all methods on the UK-Abs dataset, considering the 3 segments Background, Reasons, and Final Judgement. 
In this section, we present some more observations from these segment-wise evaluations, which could not be reported in the main paper due to lack of space.

An interesting observation is that the performances of several methods on a particular segment depend on the \textit{size} and \textit{location} of the said segment in the documents.
The FAC (Facts) segment in the In-Ext dataset and the Background segment in the UK-Abs dataset are large segments that appear at the beginning of the case documents.
On the other hand, the RPC (Ruling by Present Court) segment in In-Ext and the `Final judgement' segment in UK-Abs are short segments appearing at the end of the documents. 
Most domain-independent models, like Luhn and BERT-Ext, perform much better for the FAC and Background segments, than for the RPC and `Final judgement' segments. 
Such models may be suffering from the lead-bias problem~\cite{kedzie2018content} whereby a method has a tendency to pick initial sentences from the document for inclusion in the summary.
 
However, the RPC and `Final judgement' segments are important from a legal point of view, and should be represented well in the summary according to domain experts~\cite{bhattacharya2019comparative}.  
In fact, the performances of all methods are relatively poor for for these segments (see Table~\ref{tab:rouge-segment-india-all} and Table~\ref{tab:rouge-segment-uk-all}).
Hence, another open challenge in domain-specific long document summarization is to develop algorithms that perform well on short segments that have domain-specific importance.

\subsection{Expert Evaluation Details} \label{sec:expert-eval-details}

We mention below some more details of the expert evaluation, which could not be accommodated in the main paper due to lack of space.

\vspace{2mm}
\noindent {\bf Choice of documents for the survey:} 
We selected 5 documents from the IN-Abs test set, specifically, those five documents that gave the best average ROUGE-L F-scores over the 7 summarization methods chosen for the human evaluation. 

Ideally, some summaries that obtained lower ROUGE scores should also have been included in the evaluation by the domain experts.
But the number of summaries that we could get evaluated was limited by the availability of the experts.

\vspace{2mm}
\noindent {\bf Framing the questions asked in the survey:}
We framed the set of questions (described in Section~\ref{sec:eval}) based on the parameters stated in~\cite{bhattacharya2019comparative,achieved-textsumm} about how a legal document summary should be evaluated.

\vspace{2mm}
\noindent {\bf Pearson Correlation as IAA} : The human annotators were asked to rate the summaries on a scale of 0-5, for different parameters. Here we discuss the IAA in the `Overall' parameter.
For a particular summary of a document, consider that Annotator 1 and Annotator have given scores of 2 and 3 respectively. Now, there are two choices for calculating the IAA -- 
(i)~in a regression setup, these scores denote a fairly high agreement between the annotators, (ii)~in a classification setup, if we consider each score to be a `class', then Annotator 1 has assigned a `class 2' and Annotator 2 has assigned a `class 3'; this implies a total disagreement between the two experts.
In our setting, we find the regression setup for calculating IAA more suitable than the Classification setup. Therefore we use Pearson Correlation between the expert scores as the inter-annotator agreement (IAA) measure. 
For each algorithmic summary, we calculate the correlation between the two sets of `Overall' scores. We then take the average across all the seven  `Overall' correlation scores for the seven algorithmic summaries.

\vspace{2mm}
\noindent \textbf{Computing the correlation between human judgements and the automatic metrics}: Recall that we have 5 documents for the human evaluation. For a particular algorithm, e.g. DSDR, suppose the average `Overall score given by human annotators to the summaries of the 5 documents generated by DSDR are [$h_1$, $h_2$, $h_3$, $h_4$, $h_5$], where $h_i$ denotes the average `Overall' score given by humans for the $i^{th}$ document’s summary (range [0-1]).

Suppose, the ROUGE-1 FScore of the DSDR summaries (computed with respect to the reference summaries) are [$d_1$, $d_2$, $d_3$, $d_4$, $d_5$], where $d_i$ denotes the ROUGE-1 Fscore for the $i^{th}$ document’s DSDR-generated summary (range [0-1]).

We then compute the Pearson Correlation $c_{DSDR}$  between the list of human scores and the list of  Rouge-1 Fscores for DSDR. 
We repeat the above procedure for all the 7 algorithms for a particular metric (e.g. ROUGE-1 Fscore) to get 7 $c$ values (e.g., $c_{DSDR}$, $c_{Gist}$, etc.) and then take the average of the 7 values. This gives the final correlation between ROUGE-1 Fscore and the overall scores assigned by the human evaluators.

Likewise, we compute the correlation between other automatic metrics (e.g., ROUGE-2 Fscore, BertScore) and the human-assigned overall scores.

\subsection{Ethics and limitations statement} 
All the legal documents and summaries used in the paper are publicly available data on the Web, except the reference summaries for the In-Ext dataset which were written by the Law experts whom we consulted. 
The law experts were informed of the purpose for which the annotations/surveys were being carried out, and they were provided with a mutually agreed honorarium for conducting the annotations/surveys as well as for writing the reference summaries in the IN-Ext dataset. 

The study was performed over legal documents from two countries (India and UK). While the methods presented in the paper should be applicable to legal documents of other countries as well, it is \textit{not} certain whether the reported trends in the results (e.g., relative performances of the various summarization algorithms) will generalize to legal documents of other countries.

The evaluation study by experts was conducted over a relatively small number of summaries (35) which was limited by the availability of the experts. 
Also, different Law practitioners have different preferences about summaries of case judgements. The observations presented are according to the Law practitioners we consulted, and can vary in case of other Law practitioners.
\label{sec:appendix}


\end{document}